\documentclass[10pt,twocolumn,letterpaper]{article}

\usepackage[pagenumbers]{cvpr} 
\usepackage{multirow}
\usepackage{pifont}
\usepackage{lipsum}
\usepackage{comment}

\newcommand{\paragrapht}[1]{\vspace{-10pt}\paragraph{#1}}

\definecolor{cvprblue}{rgb}{0.21,0.49,0.74}
\usepackage[pagebackref,breaklinks]{hyperref}

\def\paperID{}
\def\confName{CVPR}
\def\confYear{2025}

\title{Cross-View Completion Models are Zero-shot Correspondence Estimators}

\author{Honggyu An$^{1*}$ ~\quad Jinhyeon Kim$^{2*}$ ~\quad Seonghoon Park$^3$ \\  Jaewoo Jung$^1$ ~\quad
Jisang Han$^1$ ~\quad Sunghwan Hong$^2$ ~\quad Seungryong Kim$^{1\dagger}$ \\\\ 
$^1$ KAIST ~\quad $^2$ Korea University ~\quad $^3$ Samsung Electronics}

\begin{document}
\twocolumn[{%
\renewcommand\twocolumn[1][]{#1}%
\maketitle
\begin{center}
    \vspace{-10pt}
    \captionsetup{type=figure}
    \includegraphics[width=\linewidth]{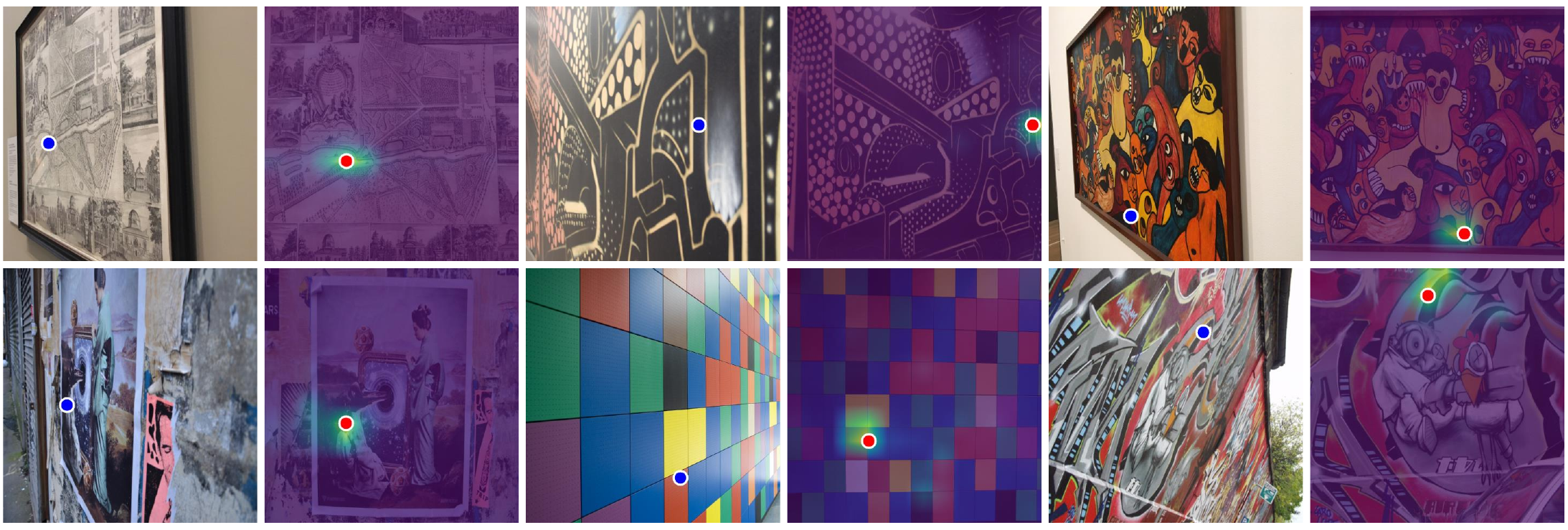}
    
    \vspace{-5pt}
    \captionof{figure}{
    \textbf{Cross-view completion models~\cite{weinzaepfel2022croco, weinzaepfel2023croco} are zero-shot correspondence estimators.} Given a pair of images consisting of target (left) and source (right) images, we visualize the attended region in the source image corresponding to a {query point} marked in the target image in \textcolor{blue}{blue}, where the point with the highest attention is marked in \textcolor{red}{red}.
    Although cross-view completion models~\cite{weinzaepfel2022croco, weinzaepfel2023croco} are not trained with correspondence-supervision, its \textbf{cross-attention} map already establishes precise correspondences.
    }
    \label{fig:teaser}
\end{center}
}]

\begingroup
\renewcommand{\thefootnote}{}
\footnotetext{$^*$: Equal contribution}
\footnotetext{$^\dagger$: Corresponding author}
\endgroup

\maketitle
\begin{abstract}

In this work, we explore new perspectives on cross-view completion learning by drawing an analogy to self-supervised correspondence learning. Through our analysis, we demonstrate that the cross-attention map within cross-view completion models captures correspondence more effectively than other correlations derived from encoder or decoder features. We verify the effectiveness of the cross-attention map by evaluating on both zero-shot matching and learning-based geometric matching and multi-frame depth estimation. 
Project page is available at \href{https://cvlab-kaist.github.io/ZeroCo/}{https://cvlab-kaist.github.io/ZeroCo/}.
\end{abstract}
    
\section{Introduction}
\label{sec:intro}
Representation learning~\citep{he2020momentum, gupta2023siamese, weinzaepfel2022croco, weinzaepfel2023croco, oquab2023dinov2} is a foundational problem in computer vision field, as it aims to extract meaningful visual cues from image collections. This enables enhanced performance in various downstream tasks such as segmentation, detection, and optical flow through transfer learning~\cite{zhu2024survey,han2024few,wang2025sea,luo2024flowdiffuser}.

Recent advances in self-supervised representation learning have introduced \textit{cross-view completion}~\cite{weinzaepfel2022croco, weinzaepfel2023croco} as a powerful pretext task, which extends the masked image modeling~\cite{he2022masked} to two views, where one view is masked and reconstructed using information from the unmasked view. Cross-view completion, in essence, leverages spatial relationships between multiple views of the same scene to learn consistent geometric features. By training models to align information across views through a cross-attention layer, this approach inherently enhances structure and geometry awareness, requiring the model to capture geometric structure and maintain spatial coherence across viewpoints. This approach has thus shown notable success, with recent works such as DUSt3R~\cite{wang2024dust3r,weinzaepfel2022croco,weinzaepfel2023croco} achieving state-of-the-art performance in complex geometric vision tasks across both 2D and 3D domains, such as optical flow, stereo depth, depth estimation, and 3D reconstruction~\cite{wang2024dust3r, weinzaepfel2023croco, revaud2024sacreg}.

However, much less is understood about the mechanisms through which the cross-view completion paradigm functions effectively. In this work, we aim to investigate and precisely identify which aspects of the representation are learned and how they contribute to improved performance on geometric downstream tasks. Our observations reveal that cross-view completion learning closely parallels \textit{self-supervised dense correspondence} learning approaches, such as optical flow estimation~\cite{liu2019selflow, jonschkowski2020mattersunsupervisedopticalflow, ren2017unsupervised, meister2018unflow, wang2019unos, shen2020ransac} and stereo depth estimation~\cite{godard2019digging, godard2017unsupervised, gordon2019depth, johnston2020self, guizilini20203d, li2021unsupervised}, as illustrated in Fig.~\ref{fig:mainqual}. Drawing from this analogy, we discover that the \textbf{cross-attention} map effectively captures geometric relationships, owing to its resemblance to conventional self-supervised dense correspondence models. 

To verify this, we analyze the learned feature representations of cross-view completion models~\cite{weinzaepfel2022croco, weinzaepfel2023croco} by visualizing pixel-wise cosine similarity scores between encoder and decoder features from input image pairs in a dense correspondence task. Specifically, we focus on CroCo-v2~\cite{weinzaepfel2023croco}, and compare similarity scores calculated from encoder features, decoder features, and the cross-attention map generated within the decoder network. As shown in Fig.~\ref{fig:attn}, these visualizations reveal that among the three correlations, the cross-attention map precisely highlights corresponding regions with greater sharpness and less noise compared to raw feature descriptors obtained directly from the encoder or decoder. Note that the encoder or decoder features of cross-view completion models have been used in previous literature for correspondence~\citep{leroy2024grounding} and 3D reconstruction~\citep{wang2024dust3r}. 

We further examine how the cross-attention map within cross-view completion models encodes a more accurate and robust representation through the lens of dense correspondence and multi-frame depth estimation. To fully leverage the learned cross-attention map for correspondence, we propose \textbf{ZeroCo}, a zero-shot inference technique that enforces reciprocity by generating a pair of cross-attention maps for the target and source images. We evaluate our approach on zero-shot dense geometric correspondence benchmarks, demonstrating that the cross-attention map encodes rich geometric representations and achieves state-of-the-art performance. Furthermore, by introducing lightweight, learnable modules on top of the cross-attention map, we achieve competitive results on standard benchmarks for dense correspondence and multi-frame depth estimation. Our extensive evaluations support these findings, showing that directly utilizing the cross-attention map significantly outperforms relying on encoder or decoder descriptors, which previous works such as DUSt3R~\cite{wang2024dust3r, leroy2024grounding} have used. Finally, we conduct comprehensive ablation studies to validate our design choices further.

In summary, our contributions are as follows:
\begin{itemize} 
\item We reveal that the effectiveness of cross-attention maps in cross-view completion models stems from their analogy to self-supervised dense correspondence learning, as their learning process closely resembles that of self-supervised correspondence learning.

\item We provide a comprehensive analysis of the learned feature representations in cross-view completion models, showing that cross-attention maps capture more accurate and robust geometric information than encoder or decoder descriptors.

\item We demonstrate that incorporating cross-attention maps with simple, lightweight heads significantly improves performance on dense correspondence and multi-frame depth estimation, achieving state-of-the-art results validated through extensive evaluations. \end{itemize}
\section{Related Work}
\label{sec:relwork}
\paragraph{Self-supervised learning.}
Self-supervised learning has proven highly effective for image-level tasks, achieving strong performance in classification and various downstream applications~\cite{caron2020unsupervised, grill2020bootstrap, he2020momentum}. Its primary advantage is the elimination of the need for ground truth data, which is hard to obtain for dense prediction tasks such as geometric matching and depth estimation. For dense prediction tasks, prior research~\cite{truong2020glu, truong2020gocor, hong2024unifying, watson2021temporal, godard2017unsupervised} has employed strategies such as generating synthetic warps or incorporating additional information, such as temporal frames and relative pose, to create supervisory signals. More recently, drawing inspiration from BERT in natural language processing~\citep{devlin2018bert}, the vision community has adopted masked image modeling~\cite{gupta2023siamese, bachmann2022multimae, he2022masked,choi2025emerging,choi2025salience} as a pretraining strategy, successfully capturing richer contextual information and finer details for fine-tuning on downstream tasks.
\vspace{-10pt}

\paragraph{Cross-view completion pretraining.}
The pretext task of cross-view completion (CVC)~\cite{gupta2023siamese, bachmann2022multimae, weinzaepfel2022croco} has garnered significant attention, enhancing robustness in geometric tasks. Extending traditional masked image modeling to two views, CVC enables the model to learn geometric knowledge by training the model to predict or align information across views. The effectiveness of CVC has been best demonstrated by CroCo-v2~\cite{weinzaepfel2023croco}, where models have achieved state-of-the-art performance in multiple downstream tasks such as visual correspondence~\citep{leroy2024grounding} and 3D reconstruction~\citep{wang2024dust3r}. While current approaches utilize CroCo-v2 by simply leveraging the decoder features, these works have not been fully exploiting the power of CVC as we reveal that the cross-attention map of the decoder embeds richer representations than the decoder features. 
\vspace{-10pt}

\paragraph{Self-supervised matching learning.}
Within the domain of dense correspondence, self-supervised learning approaches have reduced reliance on ground truth annotations by employing a variety of techniques. These techniques include photometric loss functions such as the Charbonnier penalty~\cite{yu2016back}, census loss~\cite{meister2018unflow}, and Structural Similarity Index Measure (SSIM)~\cite{wang2019unos, wang2004image}; homography transformations~\cite{truong2020glu, melekhov2019dgc, truong2020gocor}; and the incorporation of pose information~\cite{truong2023pdc, truong2021learning, edstedt2023dkm, edstedt2024roma} to establish dense correspondences. 
In self-supervised matching, source points are aligned to target points using loss functions that facilitate the refined reconstruction of the target. Here, we demonstrate that this strategy is analogous to cross-view completion pretraining, thereby enabling robust and accurate zero-shot estimation in our tasks.\vspace{-10pt}

\paragraph{Multi-frame depth estimation.} 
Monocular depth estimation has seen extensive research, maturing models~\cite{yin2023metric3d,piccinelli2024unidepth,ke2024repurposing,yang2024depth} that benefit 3D vision applications, yet lack robustness without multi-view constraints, leading to issues in complex scenes. Recent work has integrated self-supervised or geometric cues to improve robustness, with multi-frame and multi-view cost volume methods gaining popularity. Supervised~\cite{bae2022multi, facil2017single, yang2022mvs2d, li2023learning, cheng2024adaptive} and self-supervised~\cite{watson2021temporal, feng2022disentangling, guizilini2022multi, wang2023crafting, bangunharcana2023dualrefine} approaches utilize epipolar-based cost volumes for depth cues. Unlike prior works that are rely on supervised training with direct dense flow or depth regression, we show that the cross-attention map learned from cross-view completion provides readily robust cost volumes, yielding robust performance in multi-frame depth estimation tasks.
\section{Method}
\subsection{Cross-view Completion and Correspondence}
\label{sec:methods}

Cross-view completion (CVC)~\cite{weinzaepfel2022croco, weinzaepfel2023croco}, first introduced in CroCo~\cite{weinzaepfel2022croco}, is a self-supervised representation learning strategy inspired by masked image modeling~\cite{bao2021beit,he2022masked,fang2022corrupted, pathak2016context}. In CVC, the model is pretrained by reconstructing a masked target image, where about 90$\%$ of the content is obscured, using a reference source image. Specifically, features from the target and source images are initially extracted via a ViT encoder
~\cite{dosovitskiy2020image}, denoted as $D_s \in \mathbb{R}^{h \times w \times c}$ and $D_t \in \mathbb{R}^{h \times w \times c}$ from source image $I_s \in \mathbb{R}^{H \times W \times 3}$ and masked target image ${I}'_t \in \mathbb{R}^{H \times W \times 3}$ with height $h$ (or $H$), width $w$ (or $W$), and channel $c$, and then processed through a decoder containing subsequent self-attention layer, cross-attention layer, and multi-layer perceptron (MLP). In the decoder module, the cross-attention map selectively retrieves relevant information from the source features to guide the reconstruction of the target image. In this section, we draw connections between CVC and the broader self-supervised correspondence learning paradigm to highlight their analogy and shared objective of learning an effective cost volume.
\begin{figure}[!t]
    \centering
    \includegraphics[width=1.0\linewidth]{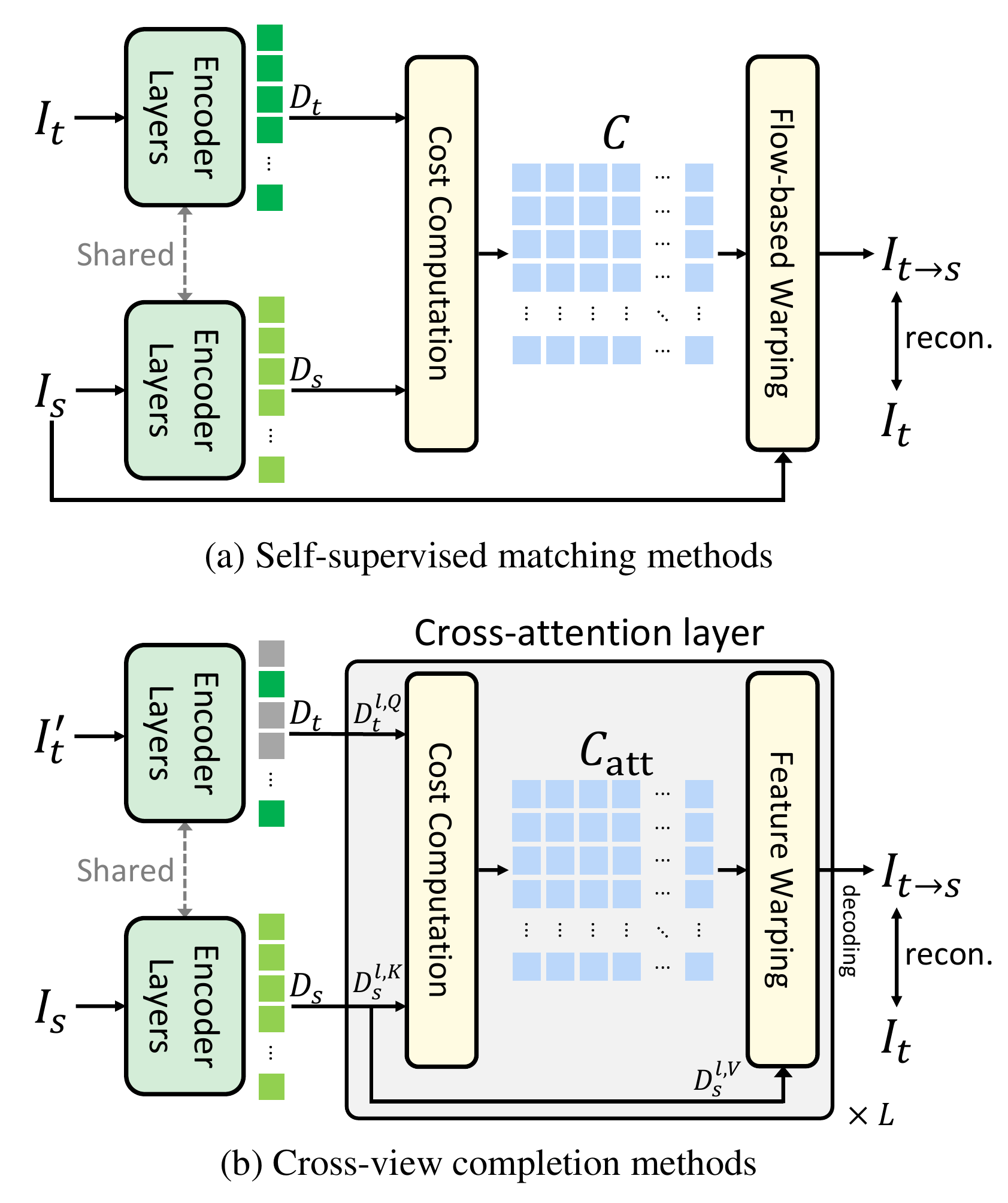}
    \vspace{-20pt}
    \caption{\textbf{Analogy of cross-view completion and self-supervised matching learning.} The cost volume learned by (b) the cross-attention layers within cross-view completion models~\cite{weinzaepfel2022croco, weinzaepfel2023croco} closely resembles that of (a) traditional self-supervised matching methods~\cite{liu2019selflow, jonschkowski2020mattersunsupervisedopticalflow}.
    }
    \vspace{-10pt}
    \label{fig:mainqual}
\end{figure}
\begin{figure*}[!t]
    \centering
    \includegraphics[width=\linewidth]{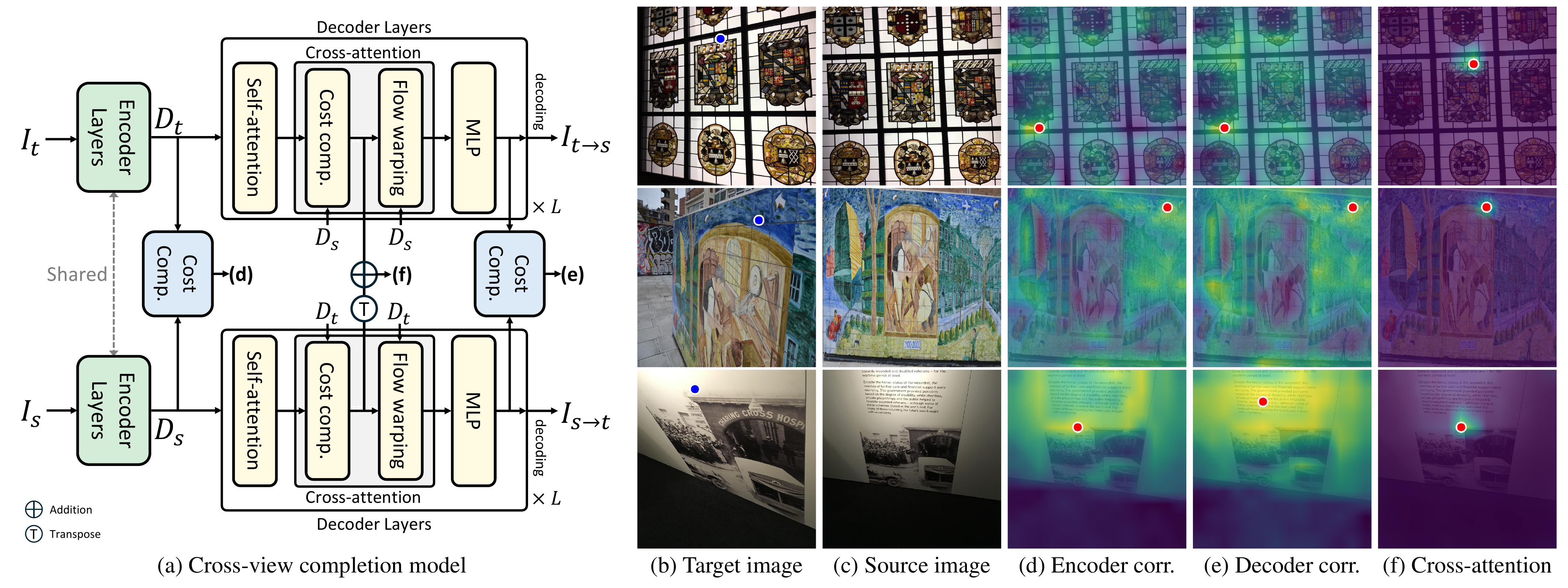}
    \vspace{-15pt}
    \caption{\textbf{Visualization of matching costs.} 
    We visualize the matching costs of the (d) encoder, (e) decoder, and (f) cross-attention maps in the (a) cross-view completion model~\cite{weinzaepfel2022croco, weinzaepfel2023croco}. The cross-attention exhibits the sharpest attention, while the encoder and decoder correlations exhibit broader attention, indicating that geometric cues are most effectively captured in the cross-attention maps.}
    \label{fig:attn}\vspace{-10pt}
\end{figure*}

\paragrapht{Cross-attention map.}
In cross-view completion, warping is guided by the cross-attention map, generated by extracting query features $D^{l,Q}_t\in \mathbb{R}^{h \times w \times d}$ with channel $d$, and key features $D_s^{l,K}\in \mathbb{R}^{h \times w \times d}$ from target $D_t$ and source $D_s$ features through projections at each $l$-th layer in decoder. The cross-attention map  $C^l_\text{att}$ is then computed by measuring the similarity between these query and key features:
\begin{equation}
    C^l_\text{att}(i,j) = \mathrm{softmax}(D_t^{l,Q}(i) \cdot D_s^{l,K}(j)/\sqrt{d}),
\label{eq:crossattn}
\end{equation}
where $l$ and $\mathrm{softmax}(\cdot)$ denote the layer index among $L$ layers and softmax operation, respectively, and $i$ and $j$ denotes pixel positions at target and source features, respectively 

\paragrapht{Cross-view completion learning.}
Based on this cross-attention map $C^l_\text{att} \in \mathbb{R}^{h \times w \times h \times w}$, the cross-attention layer warps source features into the target feature space $D^l_{t \rightarrow s}$, retrieving the source features most similar to the target:
\begin{equation}
\label{eq:crocowarp}
{D}^l_{t \rightarrow s}(i) = \sum_{j} C^l_\text{att}(i,j) \cdot  \mathrm{sampler} (D_s^{l,V};j),
\end{equation}
where $D_s^{l,V}$ represents the value features from the source image and $\mathrm{sampler}(\cdot)$ indicates a sampling operation. As the goal of CVC is to reconstruct the masked target image, an image reconstruction loss is applied to ensure that the warped features accurately reconstruct the target image:
\begin{equation}
    \mathcal{L} = \mathrm{recon}(\mathcal{E}({D}^*_{t \rightarrow s}), {I_t}),
\label{eq:crocoloss}
\end{equation}
where ${D}^*_{t \rightarrow s}$ denotes the reconstructed features from the final cross-attention layer, $\mathcal{E}(\cdot)$ is an output head that reshapes the warped features to align with the target image's dimensions, and $\mathrm{recon}(\cdot)$ indicates image reconstruction loss.

\paragrapht{Connection to self-supervised correspondence learning.}
The objective of self-supervised correspondence learning is to learn the matching networks solely with input images, $I_s$ and $I_t$, without correspondence-supervision. Correspondence methods~\cite{truong2020glu, truong2023pdc, truong2021learning,hong2022cost, hong2022neural,edstedt2023dkm, edstedt2024roma} also extract feature descriptors, $D_s$ and $D_t$, using encoders such as CNNs~\cite{he2016deep} or Transformers~\cite{vaswani2017attention}. Subsequently, a correlation map as a cost volume $C \in \mathbb{R}^{h \times w \times h \times w}$ is constructed by computing pixel-wise similarity:
\begin{equation}
C(i,j) = D_t(i) \cdot D_s(j).
\label{eq:costvolume}
\end{equation}
Following approaches in~\cite{xu2022gmflow, hong2024unifying}, this cost volume is converted into a matching distribution by applying a softmax function, thereby encoding information similar to that captured in cross-view completion pretraining~\cite{weinzaepfel2022croco}. This matching distribution identifies the most probable correspondences, which are subsequently used in loss computation to learn correspondences. The learned matching distribution ultimately enables the model to reconstruct the target image based on the information from the source image:
\begin{equation}
    {I}_{t \rightarrow s}(i) = \sum_j C(i,j) \ \mathrm{sampler}(I_s;j).
\label{eq:matchwarp}
\end{equation}
The reconstructed target image obtained in this way is then used to train the model with an reconstruction loss:
\begin{equation}
    \mathcal{L}=\mathrm{recon}({I}_{t \rightarrow s},I_t),
\label{eq:matchloss}
\end{equation}
where $\mathrm{recon}(\cdot)$ could be any reconstruction loss such as Charbonnier penalty~\cite{yu2016back} or SSIM~\cite{wang2019unos, wang2004image}.

Thus, the objective of cross-view completion pretraining closely aligns with establishing correspondences between input image pairs in a self-supervised manner, where the learning outcome is to warp the source view to reconstruct the target view. 
As shown in Fig.~\ref{fig:mainqual}, both cost volumes undergo similar computations to reconstruct the target image by learning to find the correct correspondences to minimize the matching costs between the target and source images. This indicates the potential benefits of leveraging the \textbf{cross-attention} map as a cost volume.

\paragrapht{Analysis.} Previous succesful works, such as CroCo-Flow~\cite{weinzaepfel2023croco}, DUSt3R~\cite{wang2024dust3r}, and MASt3R~\cite{leroy2024grounding}, have leveraged CVC knowledge, primarily utilizing only the \textbf{decoder} features. However, building on the analogy discussed above, we infer that the cross-attention map is trained to learn the correspondences between two input images. As shown in Fig.~\ref{fig:attn}, we visualize the attention of three correlations (d)-(f) presented in the cross-view completion model. These three correlations are derived from three components: (d) CroCo's encoder features, (e) CroCo's decoder features, and (f) CroCo's cross-attention map. Comparing the decoder feature (e) with the cross-attention map (f), we find that, surprisingly, the decoder feature-based correlation map fails to find the correct correspondences, while the cross-attention map accurately and sharply identifies the matching point. Furthermore, compared to other correlations, the cross-attention map demonstrates superior zero-shot matching performance, precisely locating the corresponding point of the blue query in the target image, while the other correlations struggle. This result indicates that the geometric knowledge learned through cross-view completion is better embodied in the cross-attention map than in the encoder and decoder features.

\subsection{Zero-shot Correspondence}
\begin{figure}[!t]
    \centering
    \includegraphics[width=1.0\linewidth]{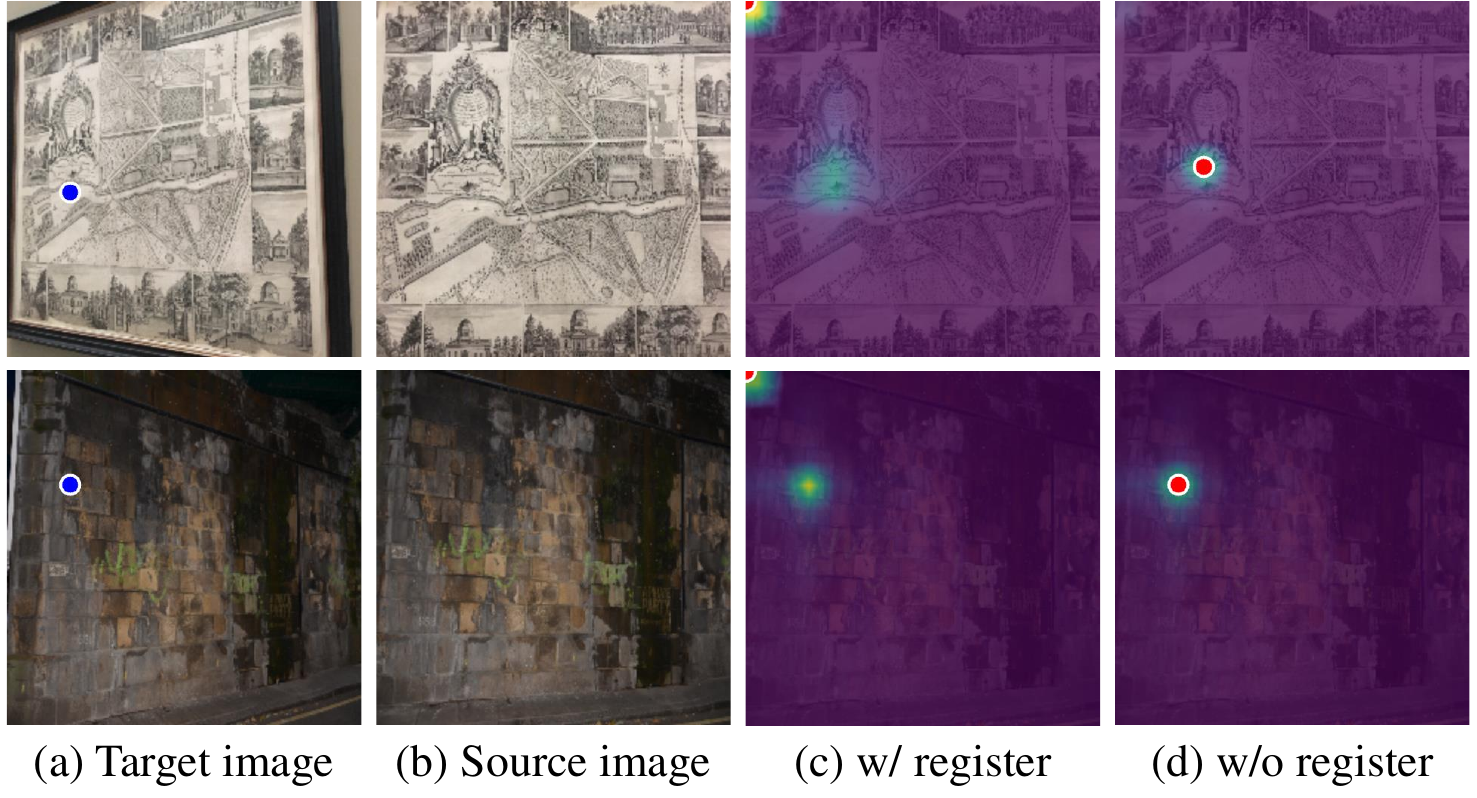}
    \vspace{-20pt}
    \caption{\textbf{Visualization of the attention map with and without the register token.} 
    The initial cross-attention map of CroCo~\cite{weinzaepfel2023croco} often contains artifacts due to the register tokens as in (c). After correcting this, the proper attending point is identified as in (d).
    }
    \vspace{-10pt}
    \label{fig:register}
\end{figure}

Based on these findings, we aim to fully utilize the learned cross-attention map for zero-shot correspondence estimation. We propose \textbf{ZeroCo}, which is a zero-shot inference technique that enforces reciprocity by generating a pair of cross-attention maps for the target and source images. Specifically, we obtain these maps by passing the original input pair $(I_t, I_s)$ and the swapped pair $(I_s, I_t)$. However, directly combining the maps is problematic due to directional constraints from the softmax operation. Therefore, we retrieve the cross-attention maps before softmax, defined as $C^l(i,j) = D_t^{l,Q}(i) \cdot D_s^{l,K}(j)$, and then combine them to enforce reciprocity. In addition, we observe that the cross-attention map $C^l$ often exhibits artifacts caused by the register token, a phenomenon linked to shortcut learning, similar to other Transformer-based models~\cite{polyak2024movie, darcet2023vision}. To address this issue, we replace the attention values of the register token with the minimum attention value. This adjustment results in more accurate attended regions in the cross-attention map, as illustrated in Fig.~\ref{fig:register}. Finally, we generate the maps $C^l$ and $C^l_\text{swap}$ from the original and swapped inputs, respectively, and fuse them as follows:

\begin{equation}
C' = \frac{1}{L}\sum_lC^l + (\frac{1}{L}\sum_lC^l_\text{swap})^T,
\end{equation}
where $C'$ denotes the final cost volume and $L$ denotes the number of cross-attention layers. From $C'$, we can infer the final flow field $F=\mathrm{softargmax}(C')$ and use this to warp the source image to the target image to evaluate the correspondences. 

\subsection{Extension: Learning-based Correspondence}
\paragraph{Geometric matching network.}
While the cross-attention map demonstrates excellent zero-shot performance, it struggles to identify fine-grained matching points due to its implicit learning nature~\cite{kim2018recurrent, rocco2018neighbourhood}. To address this, we propose two learning-based correspondence models, trained on the dataset with correspondence-supervision: one that directly fine-tunes the cross-attention map and another that utilizes an additional learnable correspondence head on the frozen cross-attention map. We name these methods \textbf{ZeroCo-finetuned} and \textbf{ZeroCo-flow}, respectively. Analogous to previous zero-shot correspondence estimation, we incorporate reciprocity into both methods by enabling the decoder to process swapped inputs and merge the resulting cross-attention maps to ensure invariance to input swapping. 

For ZeroCo-flow, we apply cost aggregation techniques to the cross-attention map such that $C' = \mathcal{T}_c(C_\text{att}, D_t) + (\mathcal{T}_c(C_\text{att,swap}, D_s))^T$, where $\mathcal{T}_c(\cdot)$ is the cost aggregation module, inspired by~\cite{cho2021cats, cho2022cats++, cho2024cat}, $C_\text{att,swap}$ represents the cross-attention map from swapped inputs, and $C'$ is the refined coarse cost volume. 


However, the resulting coarse cost volume has low resolution, which hampers the detection of fine matching details. To address this, we propose an upsampling module designed to increase resolution along the target axis, inspired by~\cite{cho2024cat}, avoiding interactions with the source axes to preserve information within the cost volume. This gives us $C'' = \mathcal{U}(C')$, where $\mathcal{U}(\cdot)$ is the upsampling module, consisting of shallow convolutional layers, and $C''$ is the upsampled cost volume. Even with these simple aggregation and upsampling modules, we achieve favorable dense geometric matching results due to the rich geometric knowledge inherent in the cross-attention map trained by cross-view completion task. The final flow is estimated as follows: $F_\mathrm{flow}=\mathrm{softargmax}(C'')$. For training, we use the same correspondence regression loss as in previous dense geometric matching works~\cite{truong2020glu, truong2021learning, truong2023pdc}.

\paragrapht{Multi-frame depth estimation network.}
In multi-frame depth estimation~\cite{watson2021temporal, guizilini2022multi}, it is common practice to construct a cost volume based on epipolar geometry to capture multi-view geometry cues that enhance depth prediction performance, which needs the accurate pose estimation, hindering the applicability and showing limited performance on dynamic regions. In this work, we aim to leverage the cross-attention map as a full cost volume for multi-frame depth estimation. However, a naive use of the cross-attention map would challenge the model in learning meaningful representations for depth estimation due to its high flexibility~\cite{kim2018recurrent, rocco2018neighbourhood}. Therefore, leveraging a cross-attention map pretrained on cross-view completion allows the model to better capture geometric cues and guide the learning process, as the cross-attention map already embodies geometric knowledge.

Similar to geometric matching, we apply cost aggregation techniques to the attention map with other features to obtain a refined cost volume. Specifically, we perform attention aggregation using both the cross-attention map and feature representations to obtain a refined cost volume, such that $C' = \mathcal{T}_d(C_\text{att},D_t)$, where $\mathcal{T}_d(\cdot)$ denotes the aggregation module for depth estimation. To further capture fine details, we design the system to produce a refined depth map $F_\mathrm{depth}$ by inputting the aggregated attention map into the DPT head~\cite{ranftl2021vision}, such that $F_\mathrm{depth} = \mathrm{DPT}(C')$. For training, we use reprojection, appearance, and smoothing losses commonly used in multi-frame self-supervised depth estimation works~\cite{watson2021temporal, godard2019digging}. Although this design is straightforward, we observe that, due to the cross-attention map's inherent rich geometric cues, our approach not only performs as well as epipolar-based cost volumes but also exhibits robustness to dynamic objects and noise. We call this model as \textbf{ZeroCo-depth}. 

The detailed architecture and aggregation technique used for ZeroCo-finetuned, ZeroCo-flow and ZeroCo-depth are provided in Sec.~\ref{supsec_imde} of the supplementary material.

\section{Experiments}
\label{sec:exp}

\begin{table*}[t!]
\centering
\resizebox{\linewidth}{!}{
\begin{tabular}{l|c|cccccc|cccccc}
\toprule
\multirow[c]{3}{*}{ Methods } &
\multirow[c]{3}{*}{ Matching cost } &
\multicolumn{6}{c|}{HPatches-240} & \multicolumn{6}{c}{HPatches-Original}\\
\cline{3-14}
& & \multicolumn{6}{c|}{AEPE$\downarrow$} & \multicolumn{6}{c}{AEPE$\downarrow$} \\
\cline{3-14}
& & \uppercase\expandafter{\romannumeral1} & \uppercase\expandafter{\romannumeral2} & \uppercase\expandafter{\romannumeral3} & \uppercase\expandafter{\romannumeral4} & \uppercase\expandafter{\romannumeral5} & Avg. & \uppercase\expandafter{\romannumeral1} & \uppercase\expandafter{\romannumeral2} & \uppercase\expandafter{\romannumeral3} & \uppercase\expandafter{\romannumeral4} & \uppercase\expandafter{\romannumeral5} & Avg. \\
\midrule \midrule

DINOv2~\cite{oquab2023dinov2} &  Correlation & 18.81 & 26.97 & 27.36 & \underline{30.66} & 36.60 & 28.08
 & 77.66& 115.11& 119.93& 133.30& 161.05& 121.41\\
DIFT$_\mathrm{SD}$~\cite{tang2023emergent} &  Correlation & 15.89
&27.08 &29.25 &32.76 &40.34 &29.06
&52.85&95.69&108.90&123.72&156.31&107.49\\
DIFT$_\mathrm{ADM}$~\cite{tang2023emergent} & Correlation & 24.21&34.88&35.92&39.62&46.76&36.28&-&-&-&-&-&- \\
DIFT$^*_\mathrm{ADM}$~\cite{tang2023emergent} & Correlation & \underline{13.57}
&\underline{24.41}
&\underline{26.16}
&31.03
&\underline{35.52}
&\underline{26.14}
&52.50& \underline{92.33}& \underline{101.60}& \underline{113.93} & \underline{143.54}& \underline{100.78}\\
SD-DINO~\cite{zhang2024tale}& Correlation & 13.98 & 26.69 & 28.79 & 34.42 & 42.04 &29.19 &\underline{48.99} &105.82 &110.49 &133.38 &157.34 &111.05\\
\midrule
CroCo~\cite{weinzaepfel2023croco} Encoder & Correlation &  39.69 &47.28 &47.35 &48.64 &54.63 &47.52 &168.56&201.52&197.82&206.63&234.83&201.87 \\
CroCo~\cite{weinzaepfel2023croco} Decoder & Correlation & 32.38&45.39&44.35&46.19&54.84&44.63&137.30&195.51&191.86&198.49&231.11&190.85 \\
CroCo~\cite{weinzaepfel2023croco} Enc. + Dec. & Correlation & 26.31&34.82&36.68&37.69& 43.71&35.84&112.39&148.92&154.97&160.06&192.07&153.68
\\
\midrule

\textbf{ZeroCo (Ours)} & Cross-attention &  \textbf{5.07}
&\textbf{7.16}
&\textbf{10.19}
&\textbf{11.37}
&\textbf{13.26}
&\textbf{9.41}&\textbf{20.75}&\textbf{27.32}&\textbf{39.10}&\textbf{43.43}&\textbf{46.35}&\textbf{35.39}\\
\bottomrule
		\end{tabular}
	}
    \vspace{-5pt}
    \caption{\textbf{Zero-shot matching results on HPatches~\cite{balntas2017hpatches}.} 
    Zero-shot performance of pretrained models is evaluated using their cost volumes on both HPatches-240 and HPatches-Original, which represent $240 \times 240$ and original resolutions, respectively. A higher scene label, such as V, corresponds to a more challenging setting with extreme geometric deformation. The best result is highlighted in \textbf{bold}, and the second best result is marked with an \underline{underline}. For the correlation map, both the CroCo Encoder and CroCo Decoder utilize all their respective encoder or decoder features. In contrast, CroCo Enc. + Dec., following CroCo-flow~\cite{weinzaepfel2023croco}, DUSt3R~\cite{wang2024dust3r}, and MASt3R~\cite{leroy2024grounding}, uses the 23rd encoder layer and the 3rd, 7th, and 11th decoder layers. $*$: adjusted hyperparameters to account for memory constraints and suboptimal performance in dense settings.
    }
    \label{tab:zeroshot_hpatches}
\end{table*}
\begin{table*}[t!]
\centering
\resizebox{0.78\linewidth}{!}{

\begin{tabular}{l|c|cccccccc}
\toprule
\multirow[c]{3}{*}{ Methods } &
\multirow[c]{3}{*}{ Matching cost } &
\multicolumn{8}{c}{ETH3D}\\
\cline{3-10}
& & \multicolumn{8}{c}{AEPE$\downarrow$}\\
\cline{3-10}
& & rate=3 & rate=5 & rate=7 & rate=9 & rate=11 & rate=13 & rate=15 & Avg. \\
\midrule \midrule
DINOv2~\cite{oquab2023dinov2} &  Correlation & 21.02 & 27.38 & 34.85 & 42.63 & 49.80 & 57.94 & 64.84 & 42.64 \\
DIFT$_\mathrm{SD}$~\cite{tang2023emergent} &  Correlation & \textbf{9.93} & \underline{12.80} & \underline{17.66} & \underline{24.05} & \underline{30.01}& \underline{38.54}& \underline{46.83}& \underline{25.69}\\
DIFT$^*_\mathrm{ADM}$~\cite{tang2023emergent} & Correlation & \underline{11.20}& 15.24 & 20.81 & 27.30 & 33.38 & 40.40 & 47.61 & 27.99 \\
SD-DINO~\cite{zhang2024tale}& Correlation &15.13 &21.43 &28.65 &36.54 &43.59 &51.48 &57.59 &31.63 \\
\midrule
CroCo~\cite{weinzaepfel2023croco} Encoder & Correlation& 57.64& 66.07& 71.78& 78.05& 81.24& 88.10& 92.78&76.52\\
CroCo~\cite{weinzaepfel2023croco} Decoder & Correlation& 30.93&  40.69& 48.58& 56.29& 60.54& 68.29& 77.76& 54.73\\
CroCo~\cite{weinzaepfel2023croco} Enc. + Dec. & Correlation& 42.15& 51.56& 56.08& 58.03& 64.41& 68.23& 75.24 & 59.39\\
\midrule
\textbf{ZeroCo (Ours)} & Cross-attention & 11.64 & \textbf{11.88} & \textbf{12.00} & \textbf{12.31} & \textbf{12.52} & \textbf{13.69} & \textbf{14.99} & \textbf{12.72} \\
\bottomrule
		\end{tabular}
	}
    \vspace{-5pt}
    \caption{\textbf{Zero-shot matching results on ETH3D~\cite{schops2017multi}.} Zero-shot performance of pretrained models by evaluating their cost volumes at the original resolutions of ETH3D. A higher scene rate, such as 15, corresponds to a more challenging setting with extreme geometric deformation. The best result is highlighted in \textbf{bold}, and the second best result is marked with an \underline{underline}. For the correlation map, both the CroCo Encoder and CroCo Decoder utilize all their respective encoder or decoder features. In contrast, CroCo Enc. + Dec., following CroCo-flow~\cite{weinzaepfel2023croco}, DUSt3R~\cite{wang2024dust3r}, and MASt3R~\cite{leroy2024grounding}, uses only the 23rd encoder layer and the 3rd, 7th, and 11th decoder layers. $*$: adjusted hyperparameters to account for memory constraints and suboptimal performance in dense settings.
    }
    \vspace{-5pt}
    \label{tab:zeroshot_eth3d}
\end{table*}
\begin{figure*}[!t]
    \centering
    \includegraphics[width=\linewidth]{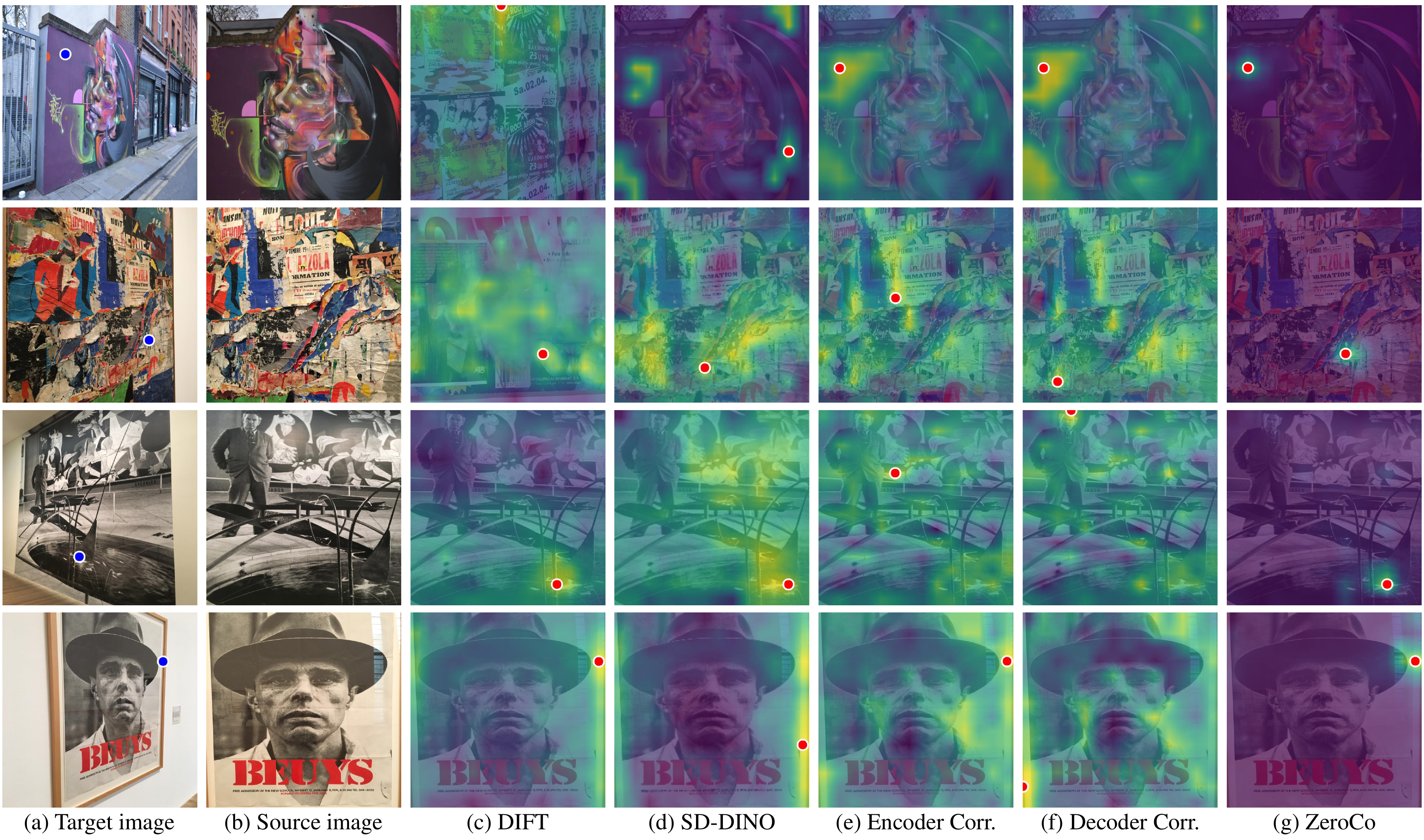}
    \vspace{-15pt}
    \caption{\textbf{Visualization of matching costs in previous zero-shot matching methods~\cite{tang2023emergent,zhang2024tale}, encoder and decoder features within cross-view completion models, and our ZeroCo.}} 
    
    \label{fig:main_attn}
\end{figure*}

\subsection{Implementation Details}
We leverage pretrained weights that were trained on cross-view completion for image reconstruction. Specifically, we adopt the pretrained weights from CroCo-v2~\cite{weinzaepfel2023croco}, where the encoder is ViT-Large~\cite{dosovitskiy2020image} and consists of 24 encoder blocks, while the decoder is made up of 12 decoder blocks. Each decoder block contains both self-attention and cross-attention layers from the original transformers~\cite{vaswani2017attention}. Further details can be found in Sec.~\ref{supsec_imde} and \ref{supsec:exde} of the supplementary material.

\begin{table*}[t!]
\centering
\resizebox{\linewidth}{!}{

\begin{tabular}{l|c|cccccc|cccccccc}
\toprule 
\multirow[c]{3}{*}{ Methods } &
\multirow[c]{3}{*}{ \shortstack{ Training \\ dataset } } &
 \multicolumn{6}{c|}{HPatches-Original} &
 \multicolumn{8}{c}{ETH3D} \\
 \cline{3-16}
& & \multicolumn{6}{c|}{AEPE$\downarrow$} & \multicolumn{8}{c}{AEPE$\downarrow$} \\
 \cline{3-16}
& & \uppercase\expandafter{\romannumeral1} & \uppercase\expandafter{\romannumeral2} & \uppercase\expandafter{\romannumeral3} & \uppercase\expandafter{\romannumeral4} & \uppercase\expandafter{\romannumeral5} & Avg. & rate=3 & rate=5 & rate=7 & rate=9 & rate=11 & rate=13 & rate=15 & Avg. \\
\midrule \midrule
DGC-Net~\cite{melekhov2019dgc} & $\mathcal{C}$ & 5.71 & 20.48 & 34.15 & 43.94 & 62.01 & 33.26 & 2.49 & 3.28 & 4.18 & 5.35 & 6.78 & 9.02 & 12.25 & 6.19 \\
GLU-Net~\cite{truong2020glu} & $\mathcal{D}$ & 1.55 & 12.66 & 27.54 & 32.04 & 52.47 & 25.05 & 1.98 & 2.54 & 3.49 & 4.24 & 5.61 & 7.55 & 10.78 & 5.17 \\
GLU-Net-GOCor~\cite{truong2020gocor} & $\mathcal{D^*}$ & \textbf{1.29} & 10.07 & 23.86 & 27.17 & 38.41 & 20.16 & 1.93 & 2.28 & 2.64 & 3.01 & 3.62 & 4.79 & 7.80 & 3.72 \\
DMP~\cite{hong2021deep} & - & 3.21 & 15.54 & 32.54 & 38.62 & 63.43 & 30.64 & 2.43 & 3.31 & 4.41 & 5.56 & 6.93 & 9.55 & 14.20 & 6.62 \\
PDCNet~\cite{truong2021learning} & $\mathcal{D^*}, \mathcal{M}$  &  \underline{1.30} & 11.92 & 28.60 & 35.97 & 42.41 & 24.04 & \underline{1.77} & 2.10 & 2.50 & 2.88 & 3.47 & 4.88 & 7.57 & 3.60 \\
PDCNet+~\cite{truong2023pdc} & $\mathcal{D^*}, \mathcal{M}$ & 1.44 & \textbf{8.97} & 22.24 & 30.13 & 31.77 & 18.91 & \textbf{1.70} & \textbf{1.96} & \textbf{2.24} & \underline{2.57} & 3.04 & 4.20 & 6.25 & 3.14 \\
DiffMatch~\cite{namdiffusion} & $\mathcal{D^*}$ & 1.85 & 10.83 & 19.18 & 26.38 & 35.96 & 18.84 & 2.08 & 2.30 & 2.59 & 2.94 & 3.29 & 3.86 & 4.54 & 3.12 \\
\midrule
DUSt3R~\cite{wang2024dust3r} & $\mathrm{MIX_8}$ & 10.11& 13.19& \textbf{15.28}& \underline{18.44}& 30.78& 17.56& 8.16& 8.93& 9.59& 10.06 & 10.77 & 11.66 & 13.57& 10.39\\
MASt3R~\cite{leroy2024grounding} & $\mathrm{MIX_{14}}$ & 13.07 & 11.58 & 20.74 & \textbf{15.18} & 24.49 & 17.01 & 1.98& 2.14& \underline{2.31} & \textbf{2.48}& \textbf{2.63} & \textbf{2.85} & \textbf{3.93} & \textbf{2.62}\\
\midrule
    \textbf{ZeroCo-finetuned (Ours)} & $\mathcal{D^*}, \mathcal{M} $ &  5.15& 11.55& 18.59& 20.86& \underline{22.72} & \underline{15.77} & 3.00& 3.49& 3.66& 4.50& 5.07&  6.01 & 7.48&  4.74\\
    \textbf{ZeroCo-flow (Ours)} &  $\mathcal{D^*}, \mathcal{M}$ & 1.51 & \underline{9.09} & \underline{15.62} & 21.07 & \textbf{20.73} & \textbf{13.61} & 1.80 & \underline{2.06} & 2.39 & \underline{2.65} & \underline{2.99} & \underline{3.60} & \underline{4.69} & \underline{2.88} \\
\bottomrule
		\end{tabular}
	}
    \vspace{-5pt}
    \caption{\textbf{Learning-based matching results on both HPatches~\cite{balntas2017hpatches} and ETH3D~\cite{schops2017multi}.} A higher scene label or rate, such as V or 15, corresponds to more challenging settings with extreme geometric deformations. The best result is highlighted in \textbf{bold}, and the second best result is marked with an \underline{underline}. The following notations are used for the training datasets: $C$: Citycam, $D$: DPED-CityScapes-ADE, $D^*$: COCO-augmented DPED-CityScapes-ADE, $\mathcal{M}$: Megadepth, $\mathrm{MIX_8}$: eight mixed dataset used in DUSt3R~\cite{wang2024dust3r}, $\mathrm{MIX_{14}}$: 14 mixed dataset used in MASt3R~\cite{leroy2024grounding}.
    }
    \vspace{-5pt}
    \label{tab:matching_main}
\end{table*}
\begin{table*}[t!]
\centering
\resizebox{0.85\textwidth}{!}{

\begin{tabular}{l|c|c|cccc|ccc}
\toprule
\multirow[c]{2}{*}{ Method } &
Additional &
Test &
\multirow[c]{2}{*}{AbsRel$\downarrow$} &
\multirow[c]{2}{*}{SqRel$\downarrow$} &
\multirow[c]{2}{*}{RMSE$\downarrow$} &
\multirow[c]{2}{*}{RMSElog$\downarrow$} &
\multirow[c]{2}{*}{$\delta_1$$\uparrow$} &
\multirow[c]{2}{*}{$\delta_2$$\uparrow$} &
\multirow[c]{2}{*}{$\delta_3$$\uparrow$}
\\
 &
network &
frames & &&&&&&
\\
\midrule
\midrule
Monodepth2      ~\citep{godard2019digging} & - &  1          & 0.115 & 0.903 & 4.863 & 0.193 & 0.877 & 0.959 & 0.981 \\
Packnet-SFM     ~\citep{guizilini20203d} & - &  1          & 0.111 & 0.785 & 4.601 & 0.189 & 0.878 & 0.960 & 0.982 \\
MonoViT         ~\citep{zhao2022monovit} & - &  1          & 0.099 & 0.708 & 4.372 & 0.175 & 0.900 & 0.967 & 0.984 \\
DualRefine      ~\citep{bangunharcana2023dualrefine} & - &  1          & 0.103 & 0.776 & 4.491 & 0.181 & 0.894 & 0.965 & 0.983 \\
GUDA            ~\citep{guizilini2021geometric} & - &  1          & 0.107 & 0.714 & 4.421 & -     & 0.883 & -     & -     \\
RA-Depth        ~\citep{he2022ra} & - &  1          & 0.096 & 0.632 & 4.216 & 0.171 & 0.903 & 0.968 & 0.985 \\
\midrule
Patil et al.    ~\citep{patil2020don} & - &  N          & 0.111 & 0.821 & 4.650 & 0.187 & 0.883 & 0.961 & 0.982 \\
ManyDepth       ~\citep{watson2021temporal} & $\mathcal{M}$ &  2 (-1, 0)  & 0.098 & 0.770 & 4.459 & 0.176 & 0.900 & 0.965 & \underline{0.983} \\
TC-Depth        ~\citep{ruhkamp2021attention} & $\mathcal{M}$ & 3 (-1, 0, 1)& 0.103 & 0.746 & 4.483 & 0.185 & 0.894 & -     & \underline{0.983} \\
DynamicDepth    ~\citep{feng2022disentangling} & $\mathcal{M}$, $\mathcal{S}$ & 2 (-1, 0)  & 0.096 & 0.720 & 4.458 & 0.175 & 0.897 & 0.964 & \textbf{0.984} \\
DepthFormer     ~\citep{guizilini2022multi} & $\mathcal{M}$ & 2 (-1, 0)  & \underline{0.090} & \underline{0.661} & \underline{4.149} & 0.175 & 0.905 & \underline{0.967} & \textbf{0.984} \\
MOVEDepth       ~\citep{wang2023crafting} & $\mathcal{M}$ &  2 (-1, 0)  & 0.094 & 0.704 & 4.389 & 0.175 & 0.902 & 0.965 & \underline{0.983} \\
DualRefine      ~\citep{bangunharcana2023dualrefine} & $\mathcal{M}$ &  2 (-1, 0)  & \textbf{0.087} & 0.698 & 4.234 & \underline{0.170} & \underline{0.914} & \underline{0.967} & \underline{0.983} \\

\midrule
\textbf{ZeroCo-depth (Ours)}       & - &  2 (-1, 0)  & \underline{0.090} & \textbf{0.637} & \textbf{4.128} & \textbf{0.169} & \textbf{0.915} & \textbf{0.968} & \textbf{0.984}\\
\bottomrule
		\end{tabular}
	}
        \vspace{-5pt}
        \caption{\textbf{Depth estimation results on the Eigen split~\cite{eigen2015predicting} of KITTI~\cite{geiger2013vision}.} We compare our model with previous single- and multi-frame depth estimation networks. The best result is highlighted in \textbf{bold}, and the second best result is marked with an \underline{underline}. $\mathcal{M}$: monocular depth network~\citep{godard2019digging}, $\mathcal{S}$: segmentation network.
    }
    \vspace{-5pt}
    \label{tab:kitti}
\end{table*}

\begin{table*}[t!]
\centering
\resizebox{0.8\textwidth}{!}{

\begin{tabular}{l|c|c|cccc|ccc}
\toprule
\multirow[c]{2}{*}{ Method } &
Additional &
Test &
\multirow[c]{2}{*}{AbsRel$\downarrow$} &
\multirow[c]{2}{*}{SqRel$\downarrow$} &
\multirow[c]{2}{*}{RMSE$\downarrow$} &
\multirow[c]{2}{*}{RMSElog$\downarrow$} &
\multirow[c]{2}{*}{$\delta_1$$\uparrow$} &
\multirow[c]{2}{*}{$\delta_2$$\uparrow$} &
\multirow[c]{2}{*}{$\delta_3$$\uparrow$}
\\
 &
networks &
frames & &&&&&&
\\
\midrule
\midrule
Monodepth2      ~\citep{godard2019digging}  & -  & 1 & 0.159 & 1.937 & 6.363 & 0.201 & 0.816 & 0.950 & 0.981 \\
ManyDepth       ~\citep{watson2021temporal}  & $\mathcal{M}$ & 2 (-1, 0) & 0.169 & 2.175 & 6.634 & 0.218 & 0.789 & 0.921 & 0.969 \\
DynamicDepth$^\dagger$    ~\citep{feng2022disentangling} & $\mathcal{M}$, $\mathcal{S}$  & 2 (-1, 0) & \underline{0.143} & \underline{1.497} & \textbf{4.971} & \underline{0.178} & \underline{0.841} & \underline{0.954} & \underline{0.983} \\
\midrule
\textbf{ZeroCo-depth (Ours)}            & - & 2 (-1, 0)  & \textbf{0.127} & \textbf{1.322} & \underline{5.058} & \textbf{0.175} & \textbf{0.860} & \textbf{0.964} & \textbf{0.987} \\
\bottomrule
		\end{tabular}
	}
    \vspace{-5pt}
    \caption{\textbf{Depth estimation results for dynamic objects in Cityscapes~\cite{cordts2016cityscapes}.}
    We compare our model with previous single- and multi-frame depth estimation networks on dynamic objects as defined in DynamicDepth~\cite{feng2022disentangling}. The best result is highlighted in \textbf{bold}, and the second best result is marked with an \underline{underline}. $\dagger$: reproduced results from the official repository. $\mathcal{M}$: monocular depth network~\citep{godard2019digging}, $\mathcal{S}$: segmentation network.
    }
    \vspace{-5pt}
    \label{tab:moving}
\end{table*}

\subsection{Evaluation Datasets and Metrics}
\paragraph{Dense geometric matching.}
To validate the effectiveness of our zero-shot and learned geometric matching, we tested our approach on the HPatches~\cite{balntas2017hpatches} and ETH3D~\cite{schops2017multi} datasets. HPatches includes sequences of different views of the same scenes, with each sequence containing one source image and five target images, along with ground-truth optical flows. Later sequences (IV and V) present more challenging targets. ETH3D features 10 real-world 3D scene sequences with transformations beyond homographies. Following \cite{truong2020glu}, we evaluate on sampled image pairs across varying intervals, covering 600K to 1,000K correspondences per interval. Larger intervals (e.g., 13 or 15) yield more challenging pairs. As per \cite{melekhov2019dgc}, we use the Average EndPoint Error (AEPE) as our metric, computed as the average Euclidean distance between estimated and ground-truth flow fields over all valid target pixels.

\vspace{-10pt}
\paragraph{Multi-frame depth estimation.}
To assess the performance of our self-supervised multi-frame depth estimation model, we evaluated it on the KITTI~\cite{geiger2013vision} and Cityscapes~\cite{cordts2016cityscapes} datasets. For KITTI, we used the standard Eigen split~\cite{eigen2015predicting} with preprocessing in \cite{zhou2017unsupervised}, resulting in 39,810 training, 4,424 validation, and 697 test images. For Cityscapes, we used 69,731 preprocessed training images and 1,525 test images with SGM-derived disparity maps~\cite{hirschmuller2007stereo}. In both datasets, we evaluated pixels with ground-truth depth less than 80 meters, clipping predicted depths accordingly.

We used standard depth evaluation metrics~\cite{eigen2015predicting} with median scaling following \cite{zhou2017unsupervised}. Lower error metric values (AbsRel, SqRel, RMSE, RMSElog) indicate better performance, while higher accuracy metric values ($\delta_1$, $\delta_2$, $\delta_3$) are preferred. For noise robustness evaluation, we adopted metrics from RoboDepth~\cite{kong2023robodepth}, including mDEE (a combined AbsRel and $\delta_1$) and mRR, which measures degradation under noisy conditions. Additional details on the dynamic and noise settings are provided in Sec.~\ref{sup_subsec:dynamicmetric} of the supplementary material.

\subsection{Zero-shot dense geometric results}
We demonstrate that the cross-attention map in cross-view completion~\cite{weinzaepfel2023croco} encodes rich geometric information by evaluating its zero-shot dense matching performance. Our method achieves state-of-the-art results on the HPatches dataset across various resolutions as shown in Tab.~\ref{tab:zeroshot_hpatches} and outperforms existing models, such as DIFT~\cite{tang2023emergent} and SD-DINO~\cite{zhang2024tale}. We conducted inference with DINO-v2~\cite{oquab2023dinov2}, renowned for its geometric awareness in visual foundation models~\cite{el2024probing}. These models experience significant performance drops under extreme geometric changes (i.e., scenes IV and V). Fig.~\ref{fig:main_attn} presents a visualization of matching costs from previous zero-shot matching methods, along with the three components in the cross-view completion model. This suggests that CVC encodes richer geometric information compared to diffusion models~\cite{rombach2022high} or DINOv2~\cite{oquab2023dinov2}. Additionally, Table~\ref{tab:zeroshot_eth3d} shows that our model can maintain robust performance on the ETH3D~\cite{schops2017multi} dataset, which also contains scenes with extreme viewpoint differences. 

The result tables evaluate whether encoder or decoder features capture the rich geometric information in CVC. The results indicate that both feature types underperform compared to DINOv2 and diffusion models and lag behind the cross-attention map. This supports our analysis that CVC shares objective similarities with self-supervised matching models, demonstrating that the cross-attention map effectively encodes dense correspondence information. Thus, the essential geometric information resides in the cross-attention map, not in the encoder or decoder features.

\subsection{Learning-based matching results}
\paragraph{Geometric matching results.}
Given the strong zero-shot matching performance of CVC's cross-attention map, we hypothesized that providing an additional matching signal would further boost dense matching performance. Tab.~\ref{tab:matching_main} demonstrates that simply incorporating a cost aggregation and upsampling module, ZeroCo-flow achieves state-of-the-art results on the HPatches dataset~\cite{balntas2017hpatches}, outperforming existing dense matching models~\cite{namdiffusion, truong2020glu, truong2023pdc, truong2020gocor, truong2021learning, hong2021deep}. In contrast, models like DUSt3R~\cite{wang2024dust3r} and MASt3R~\cite{leroy2024grounding}, which depend on extensive pretraining and decoder descriptors, fail to fully exploit CVC's geometric information. Additionally, our model shows competitive performance against MASt3R on the ETH3D dataset~\cite{schops2017multi}, even though it was trained on a smaller dataset.
\vspace{-10pt}

\begin{table}[t!]
\centering
\resizebox{0.9\linewidth}{!}{
\begin{tabular}{l|c|c|cc}
\toprule
\multirow[c]{2}{*}{ Method } &
Additional &
Noise &
\multirow[c]{2}{*}{ mDEE $\downarrow$ } &
\multirow[c]{2}{*}{ mRR  $\uparrow$ }
\\
 &
 network &
frame &
&
 \\
\midrule
\midrule
Manydepth      ~\citep{watson2021temporal}  & $\mathcal{M} $ & 0 & 0.277 & 0.803\\
DualRefine    ~\citep{bangunharcana2023dualrefine} & $\mathcal{M} $  & 0 & 0.268 & 0.801\\
\textbf{ZeroCo-depth (Ours)}            & - & 0 & \textbf{0.118} & \textbf{0.967}\\
\midrule
Manydepth      ~\citep{watson2021temporal}  & $\mathcal{M} $ & -1 & 0.118 & 0.979 \\
DualRefine    ~\citep{bangunharcana2023dualrefine} & $\mathcal{M} $  & -1 & 0.102 & 0.983\\
\textbf{ZeroCo-depth (Ours)}             & -  & -1 &\textbf{0.100}&\textbf{0.986}\\
\midrule
Manydepth      ~\citep{watson2021temporal}  & $\mathcal{M} $ & 0, -1 & 0.262 & 0.819 \\
DualRefine    ~\citep{bangunharcana2023dualrefine} & $\mathcal{M} $ & 0, -1 & 0.265& 0.805 \\
\textbf{ZeroCo-depth (Ours)}              & - & 0, -1 & \textbf{0.161} & \textbf{0.920} \\
\bottomrule
		\end{tabular}
	}
    \vspace{-5pt}
    \caption{\textbf{Depth estimation results in a practical image noise setting on KITTI~\cite{geiger2013vision}.} We follow the evaluation protocol of Robodepth~\citep{kong2023robodepth} to assess noise robustness. We measured metrics for three different scenarios: when noise is present only in the current frame, only in the previous frame, and in both frames simultaneously. $\mathcal{M}$: monocular depth network~\citep{godard2019digging}.
    }
    \vspace{-10pt}
    \label{tab:kitt_noise}
\end{table}

\begin{table}[t!]
\centering
\resizebox{\linewidth}{!}{

\begin{tabular}{lc|c|c|c}
\toprule
& Norm.& Reciprocity & Dense zoom-in & AEPE $\downarrow$ \\

\midrule \midrule
(\textbf{\uppercase\expandafter{\romannumeral1}}) & \ding{55} & \ding{55} & \ding{55} & 10.85 \\
\midrule
(\textbf{\uppercase\expandafter{\romannumeral2}}) & \ding{55} & \ding{51} & \ding{55} & 10.24\\
(\textbf{\uppercase\expandafter{\romannumeral3}}) & L2-Norm & \ding{55} & \ding{55} & 10.66 \\
(\textbf{\uppercase\expandafter{\romannumeral4}}) & Softmax & \ding{55} & \ding{55} & 10.53 \\
(\textbf{\uppercase\expandafter{\romannumeral5}}) & L2-Norm & \ding{51} & \ding{55} & 10.34\\
(\textbf{\uppercase\expandafter{\romannumeral6}}) & Softmax & \ding{51} & \ding{55} & 10.45\\
\midrule
(\textbf{\uppercase\expandafter{\romannumeral7}}) & \ding{55} & \ding{51} & \ding{51} & \textbf{9.41}\\
\bottomrule
		\end{tabular}
	}

    \caption{\textbf{Ablation Studies on HPatches-240~\cite{balntas2017hpatches} dataset.}
    }
    \vspace{-15pt}
    \label{tab:ablation}
\end{table}

\paragraph{Multi-frame depth results.}
As shown in Tab.~\ref{tab:kitti}, epipolar-based volumes also underperform against our method. By combining our high-quality cost volume with an upsampling head, we achieved state-of-the-art results on the KITTI~\cite{geiger2013vision} dataset. This demonstrates that our approach overcomes the limitations of epipolar-based methods while leveraging the geometric information encoded in the cross-attention map for superior depth estimation accuracy. 

We hypothesized that CVC's well-trained cross-attention map could replace epipolar-based cost volumes, offering improved robustness against dynamic objects and noise. Tab.~\ref{tab:moving} shows that epipolar-based cost volumes struggle with dynamic objects, leading previous methods to rely on monocular depth estimation teachers~\cite{godard2019digging} or segmentation maps~\cite{feng2022disentangling} to mitigate this issue. Additionally, Tab.~\ref{tab:kitt_noise} highlights how noise disrupts epipolar constraints, affecting cost volume estimation. In contrast, ZeroCo-depth captures geometric relationships across all pixels, not limited to the epipolar line, thereby enhancing robustness in dynamic and noisy scenarios.
Further results on multi-frame depth estimation are provided in Sec.~\ref{sup_subsec:addi_depth} of the supplementary material.

\subsection{Ablation studies}
In ablation study, we evaluate the contribution of each component to zero-shot dense matching. Table~\ref{tab:ablation} shows the quantitative results on HPatches-240~\cite{balntas2017hpatches}. The baseline (\uppercase\expandafter{\romannumeral1}) averages cross-attention maps to form the cost volume without additional components.  (\uppercase\expandafter{\romannumeral2}) to (\uppercase\expandafter{\romannumeral7}) assess the impact of specific modifications to the baseline. Specifically, (\uppercase\expandafter{\romannumeral2}) underscores the importance of reciprocity enforcement, showing that incorporating swapped inputs is more effective than relying on a single input direction.  (\uppercase\expandafter{\romannumeral3}) to (\uppercase\expandafter{\romannumeral6}) examine the impact of attention map normalization, which improves performance without reciprocity enforcement but degrades it when reciprocity is enforced. Finally,  (\uppercase\expandafter{\romannumeral7}), incorporating the dense zoom-in approach, significantly enhances zero-shot performance, highlighting its role in improving generalization.

\section{Conclusion}
\label{sec:conclusion}
In this work, we have analyzed the aspects of cross-view completion learning, especially by bringing analogies from the self-supervised correspondence learning paradigm. Our extensive experiments and analysis demonstrate that the cross-attention map best embeds the strong correspondence information when compared to previously utilized decoder features~\cite{wang2024dust3r, weinzaepfel2023croco, revaud2024sacreg}. We believe that our findings will better facilitate the training of various geometric downstream tasks, by effectively utilizing the cross-attention map of cross-view completion model.
{
    \small
    \bibliographystyle{ieeenat_fullname}
    \bibliography{main}
}









\appendix
\twocolumn[{%
\renewcommand\twocolumn[1][]{#1}%
\begin{center}
    \textbf{\Large Cross-View Completion Models are Zero-shot Correspondence Estimators}
    \vspace{1em} \\
    \textbf{\large - Supplementary Material - }
\end{center}
}]

In this supplementary material, Section~\ref{supsec_imde} outlines the implementation of zero-shot and learning-based methods, including a pipeline overview with notations, measurements, and a description of the learnable head for geometric matching and depth estimation, supported by equations and architectural illustrations. Section~\ref{supsec:exde} details the experimental setup for these methods, highlighting the baselines and configurations for dynamic objects and noisy environments in learning-based depth estimation. Section~\ref{subsec:analysis} examines cross-view completion~\cite{weinzaepfel2023croco}, focusing on layer-wise feature performance, the impact of inference resolution, pretrained weights, and additional ablation studies. Section~\ref{subsec:quan} presents extended quantitative results for zero-shot and learning-based experiments, while Section~\ref{subsec:qual} provides supplementary qualitative results to further validate our approach. Finally, Section~\ref{subsec:limitation} discusses the limitations of our proposed method.

\section{Implementation Details}
\label{supsec_imde}

\subsection{Zero-shot Matching}
For our main CroCo~\cite{weinzaepfel2023croco} zero-shot experiment reported in Table~\ref{tab:zeroshot_hpatches}, the encoder is based on ViT-Large~\cite{dosovitskiy2020image}, consisting of 24 encoder blocks, while the decoder includes 12 decoder blocks, each incorporating both self-attention and cross-attention layers. We utilized pretrained weights from CroCo-v2~\cite{weinzaepfel2023croco} trained on images with a resolution of $224 \times 224$.

As cross-view completion models are all pretrained on a resolution of 224, we resize all input source and target images, $I_s, I_t\in\mathbb{R}^{224\times224\times3}$, to fully leverage the pretrained model's knowledge. Note that with models pretrained with different resolutions (\eg 448), inference at higher resolutions is also possible. With a patch size of 16, the image is divided into 196 patches (or tokens). We then extract three components from CroCo: the encoder features $D_s, D_t\in\mathbb{R}^{hw\times c}$, decoder features $D_{s \rightarrow t}, D_{t \rightarrow s}\in\mathbb{R}^{hw \times d}$, and the cross-attention map $C_\mathrm{att} \in\mathbb{R}^{hw\times hw}$, where $hw=14\times14=196$, $c=1024$ and $d=768$. The cross-attention map is computed by averaging across the attention heads. To generate the final flow field $F$, we compute correlations for each component and apply a soft-argmax function with a temperature of $\tau=1e-4$. Finally, $F$ is upsampled to match the original image size.

\subsection{Learning-based Geometric Matching}
We conducted two types of learning-based geometric matching experiments, as described in our main paper: \textbf{ZeroCo-finetuned} and \textbf{ZeroCo-flow}. Both methods are based on the same network architecture used for zero-shot matching. In ZeroCo-finetuned, the cross-attention map is learned directly, whereas in ZeroCo-flow, the cross-attention map is further refined by a learnable head. The head used in our learning-based geometric matching comprises an aggregation module and an upsampling module, as shown in Fig.~\ref{supfig:main_architecture}. The number of cross-attention layers is $L=12$, and the aggregation module, repeated $N_a=4$ times, employs Swin-Transformer~\cite{liu2021swin} and vanilla Transformer~\cite{vaswani2017attention} blocks to aggregate the cross-attention map $C_\mathrm{att}\in\mathbb{R}^{hw\times hw}$ with the compressed decoder feature $D_{t \rightarrow s}'\in\mathbb{R}^{hw\times d'}$, where $hw=196$ and $d'=128$. In the upsampling module, the 16-th and 8-th encoder features $D_t^{16}, D_t^8 \in\mathbb{R}^{hw\times c}$ with $c=1024$ are used to guide the upsampling process, producing the final flow field $F_\text{flow}\in\mathbb{R}^{224\times224\times2}$. For ZeroCo-finetuned, we begin by averaging the $L$ stacked cross-attention maps and combining them in a reciprocal manner. Next, we apply the soft-argmax operator~\cite{lee2019sfnet} to transform them into a dense flow field, which is then upsampled to produce the final flow field $F_\text{flow} \in \mathbb{R}^{224 \times 224 \times 2}$.

\paragrapht{Flow head details.}
Our geometric matching head consists of two modules: an aggregation module and an upsampling module. The aggregation module aggregates the cross-attention map with the decoder features, while the upsampling module enhances the geometric matching performance by upsampling low-resolution features.

The aggregation module refines inaccurate matches in the cross-attention map $C_\mathrm{att}$ by leveraging the decoder feature $D_{t \rightarrow s}\in\mathbb{R}^{hw \times d}$ as guidance. First, we compress the decoder feature using linear projection and concatenate it with the cross-attention map. This combined representation is then processed by Swin-Transformer and vanilla Transformer blocks for spatial and multi-layer aggregation, respectively. The final refined cross-attention map $C'\in\mathbb{R}^{hw \times hw}$ is then obtained with reciprocal summation:
$$
C'=\mathcal{T}_c(C_\mathrm{att}, \mathcal{P}(D_{t \rightarrow s}))+(\mathcal{T}_c(C_\mathrm{att,swap}, \mathcal{P}(D_{s \rightarrow t})))^T
$$
where $\mathcal{T}_c$ denotes the successive transformer blocks for aggregation, $C_\mathrm{att,swap}$ represents the cross-attention map from swapped inputs is and $\mathcal{P}(\cdot)$ indicates linear projection, with $h=14$, $w=14$, $d=768$ and $d'=128$.

The map is progressively upsampled to $C'_\mathrm{up} \in \mathbb{R}^{2hw \times 2hw}$ and $C'' \in \mathbb{R}^{4hw \times 4hw}$ through two upsampling decoder layers, with each layer guided by the previous e
coder features $D_t^{16}$ and $D_t^8$, respectively. This process is illustrated as follows:
$$C'_\mathrm{up}=\mathcal{U}^1(C', D_t^{16})$$
$$C''=\mathcal{U}^2(C'_\mathrm{up}, D_t^{8})$$
Where $\mathcal{U}^1$ and $\mathcal{U}^2$ both comprise of multiple deconvolutional layers. The final flow field $F_\mathrm{flow}\in\mathbb{R}^{224 \times 224 \times 2}$ is obtained by applying soft-argmax followed by bilinear interpolation to $C''$ as the following:
$$F_\mathrm{flow}=\mathrm{softargmax}(C'').$$

\paragrapht{Training details.}
We adopt a two-stage training scheme, similar to conventional geometric matching works~\cite{truong2020glu, truong2021learning, namdiffusion, hong2024unifying}. In the first stage, we train on the DPED-CityScape-ADE dataset used in \cite{truong2020glu}. In the second stage, we extend the training by including the same dataset, augmented with COCO objects~\cite{truong2023pdc} and MegaDepth~\cite{li2018megadepth}, enhancing generalizability to real-world scenarios. The model is optimized using the AdamW optimizer~\cite{kingma2014adam} and trained the model for 150 epochs in the first stage with a learning rate of $1e-4$. In the second stage, the model was trained for 50 epochs with a learning rate of $5e-5$. We applied multistep learning rate decay~\cite{paszke2017automatic} in both stages and optimized only the learnable head while freezing all other components of the network.
 
\subsection{Learning-based Multi-frame Depth Estimation}
\begin{figure*}[!t]
    \centering
    \includegraphics[width=\linewidth]{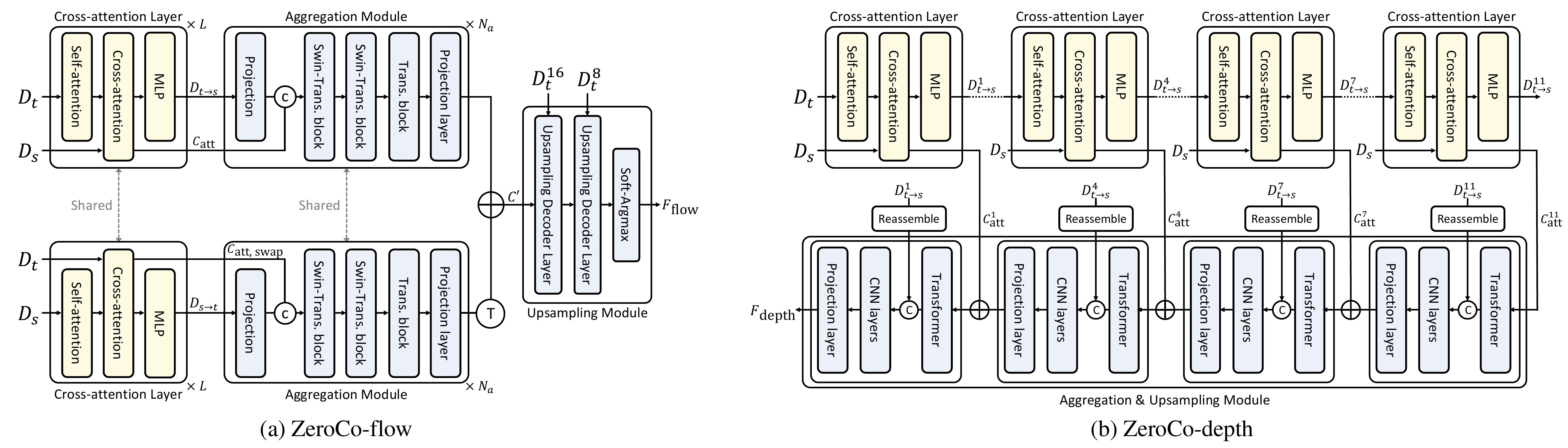}
    \vspace{-20pt}
    \caption{\textbf{Main architecture for our learning-based experiments.} (a) ZeroCo-flow and (b) ZeroCo-depth architectures extend the original zero-shot architecture by incorporating learnable heads, which consist of an aggregation and upsampling module, to effectively aggregate and refine the cross-attention map.} 
    \vspace{-10pt}
    
    \label{supfig:main_architecture}
\end{figure*}

For multi-frame depth estimation, the model receives two resized source and target images, as $I_{s}, I_{t} \in\mathbb{R}^{192 \times 640 \times 3}$. Here, the source and target images correspond to the $t$-th and $t-1$-th frames, respectively, sampled from a driving scene video. These frames are patchified (with a patch size of 16) and encoded by a transformer~\cite{dosovitskiy2020image}, resulting in encoded features $D_s, D_t \in \mathbb{R}^{h \times w \times d}$, which are then processed by a cross-attention layer to generate the cross-attention map $C_\mathrm{att} \in \mathbb{R}^{hw \times hw}$, where $h=12$, $w=40$ and $d=768$. The cross-attention map is obtained by averaging the attention heads. Similar to DPT~\cite{ranftl2021vision}, our learnable head employed for depth estimation receives four decoder features $D_{t \rightarrow s}^{l}$, and four cross-attention maps $C_\mathrm{att}^{l}$, where $l=1,4,7,11$. Its objective is to aggregate the cross-attention map, guided by the decoder features, and upsample the depth map to capture fine details. Aggregation is performed by the transformer, while upsampling is performed by CNN layers as shown in Fig.~\ref{supfig:main_architecture}. The final projection layer of our depth head outputs a depth map $F_\mathrm{depth}\in\mathbb{R}^{96 \times 320 \times 1}$. 

\paragrapht{Depth head details.}
Analogous to previous geometric matching head, our depth head consists of two modules: the aggregation module and the upsampling module. The aggregation module aggregates the cross-attention map with the decoder features, while the upsampling module upsamples the features to capture fine details of the depth map.

The aggregation module aims to refine inaccurate matches in the cross-attention map and reduces its degree of freedom. Using a transformer~\citep{vaswani2017attention}, we first compress the initial cross-attention map $C_{\mathrm{att}}$ to obtain $C'_{\mathrm{att}} \in \mathbb{R}^{h \times w \times c}$ as the following: $$C'_{\mathrm{att}} = \mathcal{T}_{\mathrm{att}}(C_{\mathrm{att}}),$$ where $\mathcal{T}_{\mathrm{att}}(\cdot)$ denotes attention aggregation. Then we use the compressed target feature $D_{t \rightarrow s}' \in \mathbb{R}^{h \times w \times d'}$ to further refine the compressed map as the following:
$$C' = \mathcal{T}_{\mathrm{feat}}([C'_{\mathrm{att}}, D_{t \rightarrow s}']), $$ where $C' \in \mathbb{R}^{h \times w \times hw}$ indicates the refined cross-attention map and $\mathcal{T}_{\mathrm{feat}}(\cdot)$ denotes the feature aggregation component composed of multiple convolutional layers. This whole process can be summarized as the following:
$$C'=\mathcal{T}_d(C_\mathrm{att},D_t) = \mathcal{T}_\mathrm{feat}([\mathcal{T}_\mathrm{att}(C_\mathrm{att}),D_{t \rightarrow s}']),$$
where $\mathcal{T}_d(\cdot)$ denotes the aggregation module for depth estimation.

Next, the upsampling module aims to mitigate the limitations of low-resolution attention maps in fine depth prediction. Thus, we stack the aggregation modules in a pyramidal structure to hierarchically refine the cost volumes, similar to coarse-to-fine feature enhancement methods~\cite{min2021hypercorrelation}. As illustrated in Fig.~\ref{supfig:main_architecture}, we adopt a four-stage pyramidal process using aggregation modules~\cite{ranftl2021vision}, selecting four layers of decoder features $D_{t \rightarrow s}^{l}$, and cross-attention maps $C_\mathrm{att}^{l}$, where $l=1,4,7,11$. Also, resolutions are progressively adjusted using the Reassemble operation~\citep{ranftl2021vision} and bilinear interpolation, enabling hierarchical feature and attention map aggregation into a refined cost volume:
$$
C_\mathrm{ref}^l = \mathcal{T}_d([\mathrm{interp}(C_{\mathrm{att}}^l + C_\mathrm{ref}^{l-1})), D_{t \rightarrow s,\mathrm{re}}^l])
$$
$$D_{t \rightarrow s, \mathrm{re}}^l = \mathrm{Reassemble}_l(D_{t \rightarrow s}^l),$$
where $l$ denotes the stage number, $\mathrm{interp}$ refers to bilinear interpolation, $\mathrm{Reassemble}_l$ indicates the Reassemble blocks for stage $l$, and $C_{\mathrm{ref},l}$ represents the refined cost volume.
The final refined cost volume input to the depth head has dimensions of $C_\mathrm{ref}^*\in\mathbb{R}^{4h \times 4w \times c}$, resulting in a final depth map $F_\mathrm{depth}\in\mathbb{R}^{96 \times 320 \times 1}$ as the following:
$$F_\mathrm{depth} = \mathcal{T}_\mathrm{depth}(C_\mathrm{ref}^*),$$
where $\mathcal{T}_\mathrm{depth}$ comprises multiple convolution layers and a final sigmoid layer. 

\paragrapht{Training details.}
We adopt the same augmentation scheme of previous depth estimation methods~\cite{godard2019digging}, our model is trained with input resolutions of $192\times640$ for the KITTI dataset~\cite{geiger2013vision} and $128 \times 416$ for the Cityscapes dataset~\cite{cordts2016cityscapes}. Experiments were conducted using PyTorch~\citep{paszke2019pytorch} on an RTX 3090 GPU with a batch size of 8. We employ the Adam~\citep{kingma2014adam} optimizer, setting the learning rate as $5e-6$ for the pretrained encoder and decoder, and $5e-5$ for the remaining components. We incorporate static augmentation from ManyDepth~\citep{watson2021temporal} with a $50\%$ ratio. For the photometric loss function, we set $\lambda=0.001$ and $\alpha=0.85$. Lastly, we employ the same pose network used in previous depth estimation methods~\cite{godard2019digging}.

\section{Experimental Details}
\label{supsec:exde}

\subsection{Zero-shot Matching}
\paragraph{Baselines.}
We utilized DINOv2~\cite{oquab2023dinov2}, DIFT$_\mathrm{ad}$~\cite{tang2023emergent}, DIFT$_\mathrm{sdm}$~\cite{tang2023emergent}, and SD-DINO~\cite{zhang2024tale} as baseline models for comparison, all trained in a self-supervised manner without labels. For DINOv2, we used features extracted from the 11-th layer to construct matching costs, which are critical for achieving optimal performance.
DIFT~\cite{tang2023emergent} employs a diffusion-based approach for zero-shot matching and is evaluated in two variants: DIFT$_\mathrm{ad}$ and DIFT$_\mathrm{sdm}$. Both variants used the hyperparameter configurations originally proposed for DIFT, however, DIFT$_\mathrm{sdm}$ exhibited instability at the original resolution due to its feature resolution being reduced by half. To enhance performance in dense matching, we adjusted the hyperparameters by setting the time step to 41, utilizing features from the 4-th layer, and employing an ensemble size of 2, which we denote as DIFT$^*_\mathrm{sdm}$. For the HPatches-240 dataset, features were extracted at a resolution of $224 \times 224$ for dense matching, while for the HPatches-original dataset, features were extracted at $768 \times 768$.
SD-DINO~\cite{zhang2024tale}, which also integrates a diffusion model with DINOv2 for zero-shot matching, was evaluated using the same hyperparameter configurations as its original implementation.

We applied L2-normalization to the features extracted from each image and constructed a correlation map to compute the matching cost. Following this, in alignment with our experimental setup, we applied a soft-argmax function with a temperature of $\tau = 1e-4$ to obtain the final flow field $F$, which was then upsampled to match the original image size.

\subsection{Learning-based Geometric Matching}
\paragraph{Baselines.}
To ensure a fair comparison, we trained a dense matching model and evaluated it on the HPatches~\cite{balntas2017hpatches} and ETH3D~\cite{schops2017multi} datasets, benchmarking against state-of-the-art models such as DiffMatch~\cite{namdiffusion}. Many recent approaches, including GLU-Net~\cite{truong2020glu}, GoCor~\cite{truong2020gocor}, and DiffMatch~\cite{namdiffusion}, were trained on DPED-Cityscapes-ADE dataset, while PDC-Net~\cite{truong2021learning} and PDC-Net+~\cite{truong2023pdc} extended their training to the MegaDepth~\cite{li2018megadepth} dataset, leveraging uncertainty modeling to address significant occlusions.

Additionally, we compared our results to models that implicitly or explicitly learn matching, such as CroCo-Flow~\cite{weinzaepfel2023croco}, DUSt3R~\cite{wang2024dust3r}, and MASt3R~\cite{leroy2024grounding} which are fine-tuned from the pretrained CroCo-v2~\cite{weinzaepfel2023croco}. CroCo-Flow was evaluated using DPED-Cityscapes-ADE dataset, while DUSt3R and MASt3R were tested using pretrained models trained on more extensive datasets, including MegaDepth. DUSt3R generates aligned point maps, enabling the use of point-cloud nearest neighbors (NN) to identify matching points, whereas MASt3R employs feature-based NN matching. For all evaluations, we performed dense matching without incorporating uncertainty or cycle-consistency checks during inference.

\subsection{Learning-based Multi-frame Depth Estimation}
\paragraph{Baselines.}
To validate our performance, we conducted comparisons with models trained in a self-supervised manner using the same splits on KITTI~\cite{geiger2013vision} and Cityscapes~\cite{cordts2016cityscapes} datasets. For monocular input, we evaluated against Monodepth2~\cite{godard2019digging}, Packnet-SFM~\cite{guizilini20203d}, MonoViT~\cite{zhao2022monovit}, GUDA~\cite{guizilini2021geometric}, and RA-Depth~\cite{he2022ra}. Among multi-input models, we included those leveraging epipolar-based cost volumes to incorporate multi-view geometric information, such as ManyDepth~\cite{watson2021temporal}, DynamicDepth~\cite{feng2022disentangling}, DepthFormer~\cite{guizilini2022multi}, MOVEDepth~\cite{wang2023crafting}, and DualRefine~\cite{bangunharcana2023dualrefine}, as well as attention-based approaches like Patil et al.~\cite{patil2020don} and TC-Depth~\cite{ruhkamp2021attention}. Notably, most multi-frame depth estimation models struggle to handle dynamic scenes and often rely on knowledge distillation from Monodepth2~\cite{godard2019digging} to address these challenges.

\paragrapht{Experiments setting on dynamic objects.}
\label{sup_subsec:dynamicmetric}
To evaluate our model's performance on dynamic objects, we followed the evaluation protocol of DynamicDepth~\cite{feng2022disentangling}. Dynamic objects were identified using an off-the-shelf semantic segmentation model, EfficientPS~\cite{mohan2021efficientps}. Unlike instance-level masks or inter-frame correspondences, all dynamic category pixels were projected together in a single step. Specifically, we segmented and measured metrics within the regions corresponding to dynamic categories, including Vehicles, Persons, and Bikes.

\paragrapht{Experiments setting on noisy environments.}
\label{sup_subsec:noisemetric}
Building on the methodology of RoboDepth~\citep{kong2023robodepth}, which evaluated depth prediction performance in noisy, real-world scenarios, we also assessed the performance of multi-frame depth estimation under challenging conditions where epipolar-based cost volumes tend to struggle. We employed their evaluation metrics, including mDEE, a combined metric of AbsRel and $\delta_1$, as well as mRR, which measures the degradation in performance relative to noise-free conditions. These metrics are defined as follows:
\begin{equation}
\text{DEE} = \frac{\text{AbsRel} - \delta_1 + 1}{2}
\end{equation}

\begin{equation}
\text{mDEE} = \frac{1}{N \cdot L} \sum_{i=1}^N \sum_{l=1}^L \text{DEE}_{i,l}
\end{equation}

\begin{equation}
\text{RR}_i = \frac{\sum_{l=1}^L(1-\text{DEE}_{i,l})}{L \times (1- \text{DEE}_\text{clean})}
\end{equation}

\begin{equation}
\text{mRR} = \frac{1}{N} \sum_{i=1}^N  \text{RR}_{i}
\end{equation}

To demonstrate the noise robustness of our framework, we evaluated it across five levels of severity using augmentations from the 'Sensor \& Movement' and 'Data \& Processing' categories. These augmentations included defocus blur, glass blur, motion blur, zoom blur, elastic transformation, color quantization, gaussian noise, impulse noise, shot noise, ISO noise, pixelate, and JPEG compression.

\begin{table*}[t!]
\centering
\resizebox{\linewidth}{!}{
\begin{tabular}{l|c|c|cccccc|cccccc}
\toprule
\multirow[c]{3}{*}{ Methods } &
\multirow[c]{3}{*}{ Matching cost } &
\multirow[c]{3}{*}{ Layers } &
\multicolumn{6}{c|}{HPatches-240} & \multicolumn{6}{c}{HPatches-Original}\\
\cline{4-15}
& & & \multicolumn{6}{c|}{AEPE$\downarrow$} & \multicolumn{6}{c}{AEPE$\downarrow$} \\
\cline{4-15}
& & & \uppercase\expandafter{\romannumeral1} & \uppercase\expandafter{\romannumeral2} & \uppercase\expandafter{\romannumeral3} & \uppercase\expandafter{\romannumeral4} & \uppercase\expandafter{\romannumeral5} & Avg. & \uppercase\expandafter{\romannumeral1} & \uppercase\expandafter{\romannumeral2} & \uppercase\expandafter{\romannumeral3} & \uppercase\expandafter{\romannumeral4} & \uppercase\expandafter{\romannumeral5} & Avg. \\
\midrule \midrule

\multirow[c]{12}{*}{ \shortstack{CroCo~\cite{weinzaepfel2023croco} \\ Encoder} } & 
\multirow[c]{12}{*}{ Correlation } & 0 &77.68 & 81.24& 80.58& 81.38& 84.60& 81.10 &331.33&347.28&348.05&354.11&363.28&348.81\\
&& 2 & 66.89 &71.31& 70.97& 72.21& 76.98& 71.67&282.28&304.07&298.76&305.63&321.48&302.44\\
&& 4 & 59.57 & 64.55& 64.22&66.41& 71.11&65.17&250.96&272.93&271.94&283.00&301.25&276.02\\
&& 6 & 55.89 & 61.87& 60.49&63.23&68.66&62.03&234.22&260.80&253.24&266.84&289.43&260.91\\
&& 8 & 49.46 & 55.76 & 54.82 & 57.03 & 62.82 &55.98&208.88&237.07&229.45&241.63&268.51&237.11\\
&& 10 & 39.07 & 46.12 & 46.35& 47.79& 54.41&46.75&167.14&198.90&194.59&205.21&234.82&200.13\\
&& 12 &34.85&41.88&42.19&44.29&50.36&42.71&149.11&181.04&176.83&189.94&216.16&182.62\\
&& 14 &30.67&37.71 & 38.56&40.83&46.45&38.84&130.57&163.24&164.71&174.61&201.05&166.83\\
&& 16 &28.73& 35.89&37.21&39.56&44.92&37.26&123.70&156.69&158.48&166.67&194.58&160.02\\
&& 18 & 27.06 & 33.35& 35.95& 37.36 & 43.91 &35.53&117.07&148.66&152.34&159.23&189.85&153.43\\
&& 20 & 26.06 &32.96&35.50&37.16&43.31&35.00&113.62&145.45&150.61&157.60&189.93&151.44\\
&& 23& 26.31 &34.82 &36.68 &37.69 &43.71 &35.84 &107.74&143.96&146.31&152.34&182.56&146.58\\
&& 0-23 & 39.69 &47.28 &47.35 &48.64 &54.63 &47.52 &168.56&201.52&197.82&206.63&234.83&201.87\\
\midrule
\multirow[c]{12}{*}{ \shortstack{CroCo~\cite{weinzaepfel2023croco} \\ Decoder} } & 
\multirow[c]{12}{*}{ Correlation } & 0 & 24.03&34.73&36.39&38.67&46.46&36.05&106.09&153.44&154.79&166.91&202.01&156.65\\
&& 1 & 22.82 & 35.18 & 35.82& 40.43 &40.30&36.73&99.02&153.25&154.57&175.31&215.75&159.58\\
&& 2 &20.84& 33.43& 33.72& 39.69& 46.14& 34.96&100.20&154.72&153.14&165.12&207.62&160.16\\
&& 3 &22.72&36.93&36.54&42.39&53.36&38.39&98.65&160.79&157.14&186.83&236.06&167.89\\
&& 4 &29.85&45.03&43.23&50.54&62.20&46.17&128.00&196.18&189.83&216.90&275.98&201.38\\
&& 5 &37.31&53.51&51.08&57.05&69.58&53.71&157.78&229.95&219.40&246.55&297.62&230.26\\
&& 6 &50.82&64.64&63.37&60.65&69.90&61.87&212.22&271.83&270.34&257.03&290.88&260.46\\
&& 7 &63.51&79.03&76.00&74.26&81.53&74.86&270.40&335.62&324.02&311.68&339.78&316.30\\
&& 8 &60.89&75.10&75.32&71.83&81.11&72.85&260.35&319.29&318.12&304.12&334.94&307.36\\
&& 9 &64.55&76.34&76.85&72.39&80.32&74.09&277.35&327.92&323.53&305.80&337.55&314.43\\
&& 10 &67.41&78.49&78.77&74.37&80.95&76.00&285.27&329.35&329.48&314.91&340.35&319.87\\
&& 11 &66.89&75.14&72.98&73.67&77.91&73.32&280.18&311.57&305.92&306.77&323.13&305.51\\
&& 0-11 &32.38&45.39&44.35&46.19&54.84&44.63&137.30&195.51&191.86&198.49&231.11&190.85\\
\midrule
\multirow[c]{2}{*}{ \shortstack{CroCo~\cite{weinzaepfel2023croco} \\ Enc. + Dec.} } & 
\multirow[c]{2}{*}{ Correlation } & all &35.15&45.60&44.42&45.85&51.80&44.16&149.22&187.30&186.18&194.45&224.14&188.26\\
&& 23 / 3,7,11$^\dagger$ &26.31&34.82&36.68&37.69& 43.71&35.84&112.39&148.92&154.97&160.06&192.07&153.68\\
\midrule
\multirow[c]{12}{*}{ \textbf{ZeroCo (Ours)} } & 
\multirow[c]{12}{*}{ Cross-attention } & 0 &9.83&19.68&22.37&27.82&31.11&22.16&41.419&85.40&92.59&116.39&137.12&94.58\\
&& 1 &8.16&11.84&15.91&19.36&20.37&15.13&34.47&47.99&63.39&76.27&81.17&60.66\\
&& 2 &7.70&11.28&15.32&17.61&19.04&14.19&31.33&44.89&59.52&69.76&78.65&56.83\\
&& 3 &6.12&9.06&12.21&14.44&16.52&11.67&25.12&35.70&45.97&55.33&63.65&45.15\\
&& 4 &6.19&9.81&12.87&14.29&16.42&11.92&25.14&37.76&49.62&53.97&63.51&46.00\\
&& 5 &5.44&8.70&11.97&13.02&14.95&10.82&22.65&32.94&44.49&48.13&55.32&40.71\\
&& 6 &6.42&9.14&13.40&13.42&15.87&11.65&25.18&34.58&50.56&52.51&61.38&44.84\\
&& 7 &5.18&\underline{7.13}&10.40&\underline{11.38}&\underline{12.77}&\underline{9.37}&21.29&27.47&39.43&44.01&48.13&36.06\\
&& 8 &\textbf{5.01}&\textbf{7.04}&\underline{10.24}&11.54&\textbf{12.40}&\textbf{9.24}&\textbf{20.61}&\textbf{26.85}&\textbf{38.84}&\underline{43.72}&\underline{46.96}&\underline{35.40}\\
&& 9 &6.29&8.46&11.69&12.62&14.60&10.73&25.74&34.01&47.18&52.23&58.66&43.56\\
&& 10 &5.47&7.43&10.40&11.96&13.28&9.71&22.23&29.33&40.90&46.36&52.19&38.20\\
&& 11 &31.17&39.51&40.35&46.63&50.80&41.70&124.98&162.55&165.44&193.11&216.15&172.44\\
&& 0-11 & \underline{5.07} &7.16 &\textbf{10.19} &\textbf{11.37} &13.26 &9.41&\underline{20.75}&\underline{27.32}&\underline{39.10}&\textbf{43.43}&\textbf{46.35}&\textbf{35.39}\\
\bottomrule
		\end{tabular}
	}
    \vspace{-7pt}
    \caption{\textbf{Zero-shot matching results with different layers on HPatches~\cite{balntas2017hpatches}.} The zero-shot performance of CroCo is evaluated across different layers by analyzing their cost volumes on HPatches-240 and HPatches-Original datasets, representing resolutions of 240 $\times$240 and the original resolution, respectively. The best results are indicated in \textbf{bold}, while the second-best results are marked with \underline{underline}. $\dagger$: The encoder and decoder features are utilized in the CroCo-flow~\cite{weinzaepfel2023croco}, DUSt3R~\cite{wang2024dust3r}, and MASt3R~\cite{leroy2024grounding}. 
    }
    \vspace{-5pt}
    \label{sup_tab:layer_analysis}
\end{table*}
\clearpage

\begin{table*}[t]
\centering
\resizebox{0.8\linewidth}{!}{
\begin{tabular}{l|c|c|cccccc}
\toprule
\multirow[c]{3}{*}{ Methods } &
\multirow[c]{3}{*}{ Matching cost } &
\multirow[c]{3}{*}{ \shortstack{Input \\Resolution } } & \multicolumn{6}{c}{HPatches-Original}\\
\cline{4-9}
& & &  \multicolumn{6}{c}{AEPE$\downarrow$} \\
\cline{4-9}
& & & \uppercase\expandafter{\romannumeral1} & \uppercase\expandafter{\romannumeral2} & \uppercase\expandafter{\romannumeral3} & \uppercase\expandafter{\romannumeral4} & \uppercase\expandafter{\romannumeral5} & Avg. \\
\midrule \midrule

\multirow[c]{3}{*}{ \shortstack{CroCo~\cite{weinzaepfel2023croco} Encoder} } & 
\multirow[c]{3}{*}{ Correlation } & $224 \times 224$ &107.74&\underline{143.96}&\underline{146.31}&\underline{152.34}&\underline{182.56}&\underline{146.58}\\
&&  $448 \times 448$ &133.71&174.30&173.62&189.65&213.98&177.05\\
&&  $672 \times 672$ &163.36&197.95&200.20&213.14&240.12&202.96\\
\midrule
\multirow[c]{3}{*}{ \shortstack{CroCo~\cite{weinzaepfel2023croco} Decoder} } & 
\multirow[c]{3}{*}{ Correlation } & $224 \times 224$ &106.09&153.44&154.79&166.91&202.01&156.65\\
&&  $448 \times 448$ &114.60&159.44&161.22&175.98&202.82&162.81\\
&&  $672 \times 672$ &147.90&186.95&189.64&203.14&224.58&190.44\\
\midrule
\multirow[c]{3}{*}{ \textbf{ZeroCo (Ours)} } & 
\multirow[c]{3}{*}{ Cross-attention } & $224 \times 224$ &\textbf{20.75}&\textbf{27.32}&\textbf{39.10}&\textbf{43.43}&\textbf{46.35}&\textbf{35.39}\\
&&  $448 \times 448$ &\underline{92.81}&173.12&184.96&226.13&256.65&186.73\\
&&  $672 \times 672$ &113.09&180.32&183.98&215.25&237.99&186.12\\
\bottomrule
		\end{tabular}
	}
    \vspace{-7pt}
    \caption{\textbf{Zero-shot matching results at varying input resolutions on HPatches.~\cite{balntas2017hpatches}.} The zero-shot performance of cross-view completion with different input resolutions is evaluated using cost volumes on HPatches-Original. The best results are highlighted in \textbf{bold}, while the second best results are marked with an \underline{underline}. CroCo~\cite{weinzaepfel2023croco}, trained exclusively at $224 \times 224$, achieves the best performance at this resolution. However, training on higher resolutions could potentially enhance its performance for larger input sizes.
    }
    \vspace{-5pt}
    \label{sup_tab:resolution_analysis}
\end{table*}

\begin{table*}[t!]
\centering
\resizebox{\linewidth}{!}{
\begin{tabular}{l|c|c|cccccc|cccccc}
\toprule
\multirow[c]{3}{*}{  \shortstack{Pretrained \\ weights} } &
\multirow[c]{3}{*}{  Enc. / Dec. } &
\multirow[c]{3}{*}{ Matching cost } &
\multicolumn{6}{c|}{HPatches-240} & \multicolumn{6}{c}{HPatches-Original}\\
\cline{4-15}
& & & \multicolumn{6}{c|}{AEPE$\downarrow$} & \multicolumn{6}{c}{AEPE$\downarrow$} \\
\cline{4-15}
& & & \uppercase\expandafter{\romannumeral1} & \uppercase\expandafter{\romannumeral2} & \uppercase\expandafter{\romannumeral3} & \uppercase\expandafter{\romannumeral4} & \uppercase\expandafter{\romannumeral5} & Avg. & \uppercase\expandafter{\romannumeral1} & \uppercase\expandafter{\romannumeral2} & \uppercase\expandafter{\romannumeral3} & \uppercase\expandafter{\romannumeral4} & \uppercase\expandafter{\romannumeral5} & Avg. \\
\midrule \midrule

\multirow[c]{3}{*}{ \shortstack{CroCo-v1~\cite{weinzaepfel2022croco}} } & 
\multirow[c]{3}{*}{ViT-B / ViT-S} &Encoder-Correlation &20.14&28.70&30.82&35.04&38.09&30.56&83.66&121.80&126.74&147.53&164.43&128.82\\
&& Decoder-Correlation 
&25.21&40.35&42.06&49.68&54.71&42.40&106.72&174.56&179.56&213.87&236.85&182.31 \\
&& Cross-attention &22.77&38.04&40.05&46.04&50.13&39.41&96.27&165.30&169.30&196.17&216.41&168.69\\
\midrule
\multirow[c]{3}{*}{ \shortstack{CroCo-v2~\cite{weinzaepfel2023croco}} } & 
\multirow[c]{3}{*}{ViT-B / ViT-S} &Encoder-Correlation &16.09&22.55&25.51&27.45&31.86&24.69&69.46&96.43&104.09&115.62&134.44&104.01\\
&& Decoder-Correlation 
&18.04&26.84&28.05&30.23&38.19&28.27&78.40&117.18&118.27&129.57&162.81&121.25 \\
&& Cross-attention &7.30&11.15&14.94&16.47&19.16&13.81&30.59&47.40&58.83&66.42&79.59&56.57\\
\midrule
\multirow[c]{3}{*}{ \shortstack{CroCo-v2~\cite{weinzaepfel2023croco}} } & 
\multirow[c]{3}{*}{ViT-B / ViT-B} &Encoder-Correlation &16.16&22.35&24.59&27.28&30.25&24.13&68.78&94.61&103.07&113.95&125.60&101.20\\
&& Decoder-Correlation 
&18.89&25.76&28.90&30.76&37.36&28.33&82.98&113.79&122.72&129.42&160.06&121.80 \\
&& Cross-attention &\textbf{5.89}&\underline{9.30}&\underline{13.51}&\underline{14.21}&\underline{17.27}&\underline{12.03}&\textbf{24.85}&\underline{36.24}&\underline{51.65}&\underline{58.46}&\underline{65.15}&\underline{47.27}\\
\midrule
\multirow[c]{3}{*}{ \shortstack{CroCo-v2~\cite{weinzaepfel2023croco}} } & 
\multirow[c]{3}{*}{ViT-L / ViT-B} &Encoder-Correlation &24.62&32.92&34.75&35.80&41.69&33.96&105.02&141.72&146.45&151.03&182.55&145.36 \\
&& Decoder-Correlation &22.77&33.23&34.79&37.31&44.78&34.58&98.57&145.00&145.74&161.78&193.35&148.89\\
&& Cross-attention 
&\underline{6.09}&\textbf{8.47}&\textbf{11.87}&\textbf{12.89}&\textbf{13.83}&\textbf{10.63}&\underline{25.10}&\textbf{33.03}&\textbf{45.05}&\textbf{50.26}&\textbf{54.24}&\textbf{41.54} \\

\bottomrule
		\end{tabular}
	}
    \vspace{-7pt}
    \caption{\textbf{Zero-shot matching results with different pretrained weights on HPatches~\cite{balntas2017hpatches}.} The zero-shot performance of CroCo with different pretrained weights is evaluated using cost volumes on HPatches-240 and HPatches-Original datasets, representing  $240 \times 240$ and original resolutions, respectively. The best results are highlighted in \textbf{bold}, while the second best results are marked with an \underline{underline}.
    }
    \vspace{-5pt}
    \label{sup_tab:diffweight}
\end{table*}
\begin{table*}[t!]
\centering
\resizebox{0.9\linewidth}{!}{
\begin{tabular}{l|c|c|c|cccccc}
\toprule
\multirow[c]{3}{*}{ \shortstack{Computing \\ Matching cost} } &
\multirow[c]{3}{*}{  Normalize } &
\multirow[c]{3}{*}{ Reciprocity } &
\multirow[c]{3}{*}{ \shortstack{Dense \\zoom-in} } &
\multicolumn{6}{c}{HPatches-240} \\
\cline{5-10}
& & & & \multicolumn{6}{c}{AEPE$\downarrow$} \\
\cline{5-10}
& & & & \uppercase\expandafter{\romannumeral1} & \uppercase\expandafter{\romannumeral2} & \uppercase\expandafter{\romannumeral3} & \uppercase\expandafter{\romannumeral4} & \uppercase\expandafter{\romannumeral5} & Avg. \\
\midrule \midrule

Query-Query & \ding{55} & \ding{55} & \ding{55} & 11.72 & 21.04 & 22.11 & 29.10 & 37.04 & 24.20\\
Key-Key & \ding{55} & \ding{55} & \ding{55} & 28.28 & 42.54 & 43.99 & 50.98 & 55.56 & 44.27 \\
Value-Value & \ding{55} & \ding{55} & \ding{55} & 35.00 & 45.56 & 47.47 & 49.64 & 53.84 & 46.30\\
\midrule
Query-Key & \ding{55} & \ding{55} & \ding{55} & 6.30 & 8.28 & 12.16 & 12.63 &  14.88& 10.85\\
Query-Key & L2-Norm & \ding{55} & \ding{55} & 5.90 & 8.07 & 11.65 & 12.69 & 14.98 & 10.66\\
Query-Key & Softmax & \ding{55} & \ding{55} & 6.00 & 8.34 & 11.26 & 12.42 & 14.61 &10.53\\
Query-Key & \ding{55} & \ding{51} & \ding{55} & 5.93 & 8.13 & 11.56 & 12.46 & 13.12 & 10.24\\
Query-Key & L2-Norm & \ding{51} & \ding{55} & 5.93 & 8.19 & 11.50 & 12.51 & 13.54 & 10.34\\
Query-Key & Softmax & \ding{51} & \ding{55} & 6.06 & 8.35 & 11.44 & 12.85 & 13.52 & 10.45\\
\midrule
\textbf{Query-Key (Ours)} & \ding{55} & \ding{51} & \ding{51} & \textbf{5.07} &\textbf{7.16} &\textbf{10.19} &\textbf{11.37} &\textbf{13.26} &\textbf{9.41} \\
\bottomrule
		\end{tabular}
	}
    \vspace{-5pt}
    \caption{\textbf{Ablation studies on HPatches-240~\cite{balntas2017hpatches}.} We conduct ablation studies on the HPatches-240. 
    }
    \vspace{-5pt}
    \label{sup_tab:zero_abl}
\end{table*}
\clearpage

\section{Analysis}
\label{subsec:analysis}

\subsection{Layer Analysis of CVC Models}
As with DIFT~\cite{tang2023emergent}, we evaluated the performance of CroCo's encoder, decoder, and attention map across individual layers on the HPatches~\cite{balntas2017hpatches} dataset to identify the layer yielding the best matching cost. Additionally, we assessed performance by averaging the matching costs obtained from all layers. We also tested the performance of features used in head training for models like CroCo-flow, DUSt3R, and MASt3R, specifically the 23rd encoder feature and the 3rd, 7th, and 11th decoder features, by averaging their correlations.

The results, presented in Tab.~\ref{sup_tab:layer_analysis}, reveal that the attention map consistently outperformed the encoder and decoder in matching cost across all layers. The table further highlights that encoder layers tend to produce better matching costs in later layers, while decoder layers perform better in earlier layers. In contrast, the cross-attention mechanism achieved its best performance at intermediate layers (specifically the 7th and 8th). Notably, for datasets like Hpatches-Original, averaging the matching costs across layers yielded the best results.
Furthermore, the features used for head training in existing models like CroCo-flow~\cite{weinzaepfel2023croco}, DUSt3R~\cite{wang2024dust3r}, and MASt3R~\cite{leroy2024grounding} demonstrated descriptor qualities compared to other layer features. However, they exhibited inferior performance compared to the matching cost stored in the cross-attention maps. These results suggest that cross-attention maps encode superior geometric information compared to encoder or decoder features. Consistent with the findings from our learning-based model training, incorporating cross-attention into models like DUSt3R and MASt3R would enable them to leverage the geometric information inherent in CroCo more effectively, surpassing the performance of models relying solely on feature-based approaches.
\subsection{Resolution Analysis of CVC Models}
In our experiments, we also conducted zero-shot evaluations at the input resolution of $224 \times 224$, which is the resolution used during cross-view completion training. Additionally, we plan to extend this evaluation to higher resolutions, specifically $2 \times$ and $3 \times$ scales, at $448 \times 448$ and $672 \times 672$, to measure the performance at these resolutions.

Tab.~\ref{sup_tab:resolution_analysis} indicate that increasing the resolution degrades the performance of not only the encoder-decoder correlation map but also the cross-attention map. This phenomenon is likely due to CroCo being trained exclusively at a resolution of $224 \times 224$, where the cross-attention map achieves its best performance. Consequently, we hypothesize that if CroCo were trained at higher resolutions, it would produce effective cross-attention maps even at those resolutions, leading to finer zero-shot matching results.

\subsection{Different CVC Models}

CroCo provides four types of pretrained weights, including CroCo-v1~\cite{weinzaepfel2022croco} and CroCo-v2~\cite{weinzaepfel2023croco} with different encoder-decoder sizes. We analyzed the zero-shot matching performance of these weights.

As shown in Tab.~\ref{sup_tab:diffweight}, the cross-attention maps of CroCo-v2 models consistently capture better geometric information than the encoder-decoder correlation maps. Additionally, larger model sizes demonstrated improved matching performance. Interestingly, while CroCo-v1 was trained using the same method, its cross-attention maps showed performance comparable to correlation maps. This difference likely stems from CroCo-v2's use of a retrieval-based approach with more diverse and extensive training data, enabling better learning of cross-attention maps.

\subsection{Ablation Study}
We also conducted additional ablation studies, as detailed in Tab.~\ref{sup_tab:zero_abl}, examining the impact of various factors, including the method for constructing matching costs based on cross-attention, normalization, reciprocity, and dense zoom-in.

Initially, we computed matching costs by multiplying the query-key matrices before applying softmax. However, with reciprocity-based inference, we obtain query, key, and value features for both the target and source images, allowing us to calculate correlations between them to construct matching costs. Interestingly, as shown in the results, the original query-key method yielded the best performance.
For normalization, excluding reciprocity, both L2-normalization and softmax improved performance. However, applying normalization after combining costs through reciprocity resulted in reduced performance.
Additionally, incorporating dense zoom-in provided slight performance improvements. Our final model omits normalization while utilizing reciprocity and dense zoom-in, achieving the best overall results.

\begin{table}[t!]
\centering
\resizebox{0.9\linewidth}{!}{
\begin{tabular}{l|c|c}
\toprule
Methods &
Time(ms) &
Memories(MB) \\
\midrule \midrule

DINOv2~\cite{oquab2023dinov2} & \textbf{56.74} & \textbf{336.82}\\
DIFT$_\mathrm{SD}$~\cite{tang2023emergent} & 345.91 & 4,888.75\\
DIFT$_\mathrm{ADM}$~\cite{tang2023emergent} & 174.31 & 2,118.89\\
SD-DINO~\cite{zhang2024tale} & 5,016.78 & 9,095.14\\
\midrule
\textbf{ZeroCo w/o dense zoom-in} & \underline{100.52} & \underline{1,607.81} \\
\textbf{ZeroCo} (\textbf{ZeroCo-finetuned}) & 2,648.01 & 1,608.00 \\

\bottomrule
		\end{tabular}
	}
    \vspace{-7pt}
    \caption{\textbf{Evaluation results of memory and time complexity on HPatches-240~\cite{balntas2017hpatches}.} We conduct all experiments on a single RTX 3090 GPU. The best results are highlighted in \textbf{bold}, and the second-best results are marked with \underline{underline}.
    }
    \vspace{-5pt}
    \label{sup_tab:memory_time}
\end{table}

\subsection{Time and Memory Complexity}
We also compared the time complexity and GPU memory usage of our model against other zero-shot matching models. For this analysis, we averaged the inference time and GPU memory consumption per image pair on the HPatches-240 dataset, with all models operating at a resolution of $224 \times 224$. All experiments were conducted on a single NVIDIA RTX 3090 GPU.

As shown in Tab.~\ref{sup_tab:memory_time}, diffusion-based models~\cite{tang2023emergent, zhang2024tale} exhibit high memory and time complexity due to their large model sizes. Notably, SD-DINO~\cite{zhang2024tale} requires even more resources as it combines stable diffusion and DINOv2 features. In contrast, models like DINOv2~\cite{oquab2023dinov2}, which use only a ViT encoder, benefit from faster speeds and lower memory usage but suffer from inferior performance.
Our model uses slightly more resources than DINOv2, yet it remains significantly lighter and faster than diffusion-based models while delivering superior performance.
\section{Additional Quantitative Results}
\label{subsec:quan}

\subsection{Learning-based Geometry Matching}

\begin{table*}[t!]
\centering
\resizebox{\linewidth}{!}{

\begin{tabular}{l|c|c|cccccc|cccccc}
\toprule 
\multirow[c]{3}{*}{ Methods } &
\multirow[c]{3}{*}{ \shortstack{ Training \\ dataset } } &
\multirow[c]{3}{*}{ \shortstack{ Matching \\ cost } } &
 \multicolumn{6}{c|}{HPatches-240} &
 \multicolumn{6}{c}{HPatches-Original} \\
 \cline{4-15}
& & & \multicolumn{6}{c|}{AEPE$\downarrow$} & \multicolumn{6}{c}{AEPE$\downarrow$} \\
 \cline{4-15}
& & & \uppercase\expandafter{\romannumeral1} & \uppercase\expandafter{\romannumeral2} & \uppercase\expandafter{\romannumeral3} & \uppercase\expandafter{\romannumeral4} & \uppercase\expandafter{\romannumeral5} & Avg. & \uppercase\expandafter{\romannumeral1} & \uppercase\expandafter{\romannumeral2} & \uppercase\expandafter{\romannumeral3} & \uppercase\expandafter{\romannumeral4} & \uppercase\expandafter{\romannumeral5} & Avg. \\

\midrule \midrule
DUSt3R~\cite{wang2024dust3r} Encoder & 
$\mathrm{MIX_8}$ & Correlation &18.43&27.62&28.98&32.89&40.99&29.78&79.54&123.12&123.23&137.96&170.49&126.87\\
DUSt3R~\cite{wang2024dust3r} Decoder & $\mathrm{MIX_8}$ & Correlation 
&17.01&27.42&28.03&32.49&40.14&29.02&72.22&118.66&121.22&137.65&169.66&123.88 \\
DUSt3R~\cite{wang2024dust3r} & $\mathrm{MIX_8}$ & Cross-attention &6.49&9.00&10.64&14.04&16.78&11.39&27.75&38.28&43.12&52.97&63.84&45.19\\
MASt3R~\cite{leroy2024grounding} Encoder & 
$\mathrm{MIX_{14}}$ & Correlation &24.30&31.97&33.91&35.52&42.47&33.63&105.25&133.68&141.91&149.18&178.33&131.67\\
MASt3R~\cite{leroy2024grounding} Decoder & 
$\mathrm{MIX_{14}}$ & Correlation 
&22.23&29.34&31.28&33.40&40.61&31.37&94.06&125.98&131.58&144.19&170.76&133.31 \\
MASt3R~\cite{leroy2024grounding} & 
$\mathrm{MIX_{14}}$ & Cross-attention &5.65&6.41&8.14&8.16&8.52&7.38&24.00&27.08&32.16&33.72&38.91&31.17\\

\midrule
CroCo-flow~\cite{weinzaepfel2023croco} & $\mathrm{MIX_4}$ & - & 5.21 & 19.47 & 21.70 & 21.44 & 28.82 & 19.33 & 27.33&94.37&93.51&107.60&147.43&94.05 \\
CroCo-flow~\cite{weinzaepfel2023croco} & $\mathrm{MIX_4}, \mathcal{D}$ & - & \textbf{0.43} & \underline{2.64} & 7.73 & 8.79 & 10.63 & 6.05 & \textbf{1.32}&\textbf{7.80}&24.88&30.53&38.75&20.66 \\
DUSt3R~\cite{wang2024dust3r} & $\mathrm{MIX_8}$ & Point-Corr. & 4.18 & 6.32 & 10.15 & 11.35 & 15.32 & 9.46 &10.11& 13.19& \textbf{15.28}& \underline{18.44}& 30.78& 17.56 \\
MASt3R~\cite{leroy2024grounding} & $\mathrm{MIX_{14}}$ & Feature-Corr. & 1.31 & \textbf{1.36} & \textbf{4.01} & \textbf{2.06} & \textbf{4.34} & \textbf{2.62} &13.07 & 11.58 & 20.74 & \textbf{15.18} & 24.49 & 17.01 \\
\midrule
\textbf{ZeroCo-finetuned (Ours)} & $\mathcal{D^*}, \mathcal{M}$ & Cross-attention &1.86& 4.03& 6.03& 6.63& 7.79&5.26& 5.15& 11.55& 18.59& 20.86& \underline{22.72} & \underline{15.77}\\
\textbf{ZeroCo-flow (Ours)} & $\mathcal{D^*}, \mathcal{M}$ & Refined cost &\underline{ 0.49} & 2.73 & \underline{5.46} & \underline{6.40} & \underline{6.25} & \underline{4.27} & \underline{1.51} & \underline{9.09} & \underline{15.62} & 21.07 & \textbf{20.73} & \textbf{13.61} \\
\bottomrule
		\end{tabular}
	}
    \vspace{-7pt}
    \caption{\textbf{Dense geomtric matching results on different resolutions of HPatches~\cite{balntas2017hpatches}.} A higher scene label such as V corresponds to a more challenging setting with extreme geometric deformation. The best result is highlighted in \textbf{bold}, and the second-best result is marked with \underline{underline}. The following notations are used in the table: $D$: DPED-CityScapes-ADE, $D^*$: COCO-augmented DPED-CityScapes-ADE, $\mathcal{M}$: Megadepth, $\mathrm{MIX_4}$: 4 mixed dataset used in CroCo-flow~\cite{weinzaepfel2023croco}, $\mathrm{MIX_8}$: 8 mixed dataset used in DUSt3R~\cite{wang2024dust3r}, $\mathrm{MIX_{14}}$: 14 mixed dataset used in MASt3R~\cite{leroy2024grounding}, Corr.: Correlation.
    }
    \vspace{-5pt}
    \label{suptab:matching_main}
\end{table*}

\paragraph{HPatches results.} We compared models trained on extended datasets, including CroCo-flow~\cite{weinzaepfel2023croco}, DUSt3R~\cite{wang2024dust3r}, and MASt3R~\cite{leroy2024grounding}, all leveraging pretrained CroCo. For CroCo-flow, comparisons were made between models trained on flow datasets and DPED-Cityscapes-ADE datasets, while DUSt3R and MASt3R were evaluated on their respective mixed datasets. Additionally, we analyzed the geometric understanding encoded in the encoder, decoder, and cross-attention maps of DUSt3R and MASt3R.

Tab.~\ref{suptab:matching_main} reveals that CroCo-flow, trained on the flow dataset, underperformed due to its focus on small displacements and the limited geometric understanding of its encoder and decoder, even with additional dense-matching datasets. In contrast, DUSt3R and MASt3R showed improved geometric comprehension in encoder, decoder, and cross-attention features with larger datasets. Notably, MASt3R excelled in cross-attention maps, highlighting the structural advantage of cross-attention for capturing geometric information. 

\subsection{Learning-based Depth Estimation}
\label{sup_subsec:addi_depth}
\paragraph{Cityscape results.} Alongside KITTI~\cite{geiger2013vision} results, we present Cityscapes~\cite{cordts2016cityscapes} results, showing our model's competitive performance against state-of-the-art models in Tab.~\ref{suptab:cityscape}. Unlike methods like Monodepth2~\cite{godard2019digging} that rely on additional networks or labels, our approach harnesses the geometric quality of cross-attention alone. This demonstrates that a well-constructed cost volume can achieve strong performance in multi-frame depth estimation.

\begin{table*}[t!]
\centering
\resizebox{0.85\textwidth}{!}{

\begin{tabular}{l|c|c|cccc|ccc}
\toprule
\multirow[c]{2}{*}{ Methods } &
Additional &
Test &
\multirow[c]{2}{*}{AbsRel$\downarrow$} &
\multirow[c]{2}{*}{SqRel$\downarrow$} &
\multirow[c]{2}{*}{RMSE$\downarrow$} &
\multirow[c]{2}{*}{RMSElog$\downarrow$} &
\multirow[c]{2}{*}{$\delta_1$$\uparrow$} &
\multirow[c]{2}{*}{$\delta_2$$\uparrow$} &
\multirow[c]{2}{*}{$\delta_3$$\uparrow$}
\\
 &
network &
frames & &&&&&&
\\
\midrule
\midrule
Struct2Depth 2  ~\cite{casser2019unsupervised} & - & 1 & 0.145 & 1.737 & 7.280 & 0.205 & 0.813 & 0.942 & 0.976 \\
Monodepth2      ~\cite{godard2019digging}     & - & 1 & 0.129 & 1.569 & 6.876 & 0.187 & 0.849 & 0.957 & 0.983 \\
Videos in the Wild ~\cite{gordon2019depth}    & - & 1 & 0.127 & 1.330 & 6.960 & 0.195 & 0.830 & 0.947 & 0.981 \\
Li et al. \cite{li2021unsupervised}   & - & 1 & 0.119 & 1.290 & 6.980 & 0.190 & 0.846 & 0.952 & 0.982 \\
Struct2Depth 2  ~\cite{casser2019unsupervised} & $\mathcal{M}$, $\mathcal{F}$ & 3 (-1, 0, 1) & 0.151 & 2.492 & 7.024 & 0.202 & 0.826 & 0.937 & 0.972 \\
ManyDepth       ~\cite{watson2021temporal}   & $\mathcal{M}$  & 2 (-1, 0) & 0.114 & 1.193 & 6.223 & 0.170 & \underline{0.875} & 0.967 & \underline{0.989} \\
DynamicDepth$^\dagger$   ~\cite{feng2022disentangling} & $\mathcal{M}$, $\mathcal{S}$ & 2 (-1, 0) & \textbf{0.104} & \textbf{1.009} & \textbf{5.991} & \textbf{0.150} & \textbf{0.889} & \underline{0.972} & \textbf{0.991} \\
\midrule
\textbf{ZeroCo-depth (Ours)}                 & - & 2 (-1, 0) & \underline{0.105} & \underline{1.050} & \underline{6.117} & \underline{0.162} & \textbf{0.889} & \textbf{0.973} & \textbf{0.991} \\
\bottomrule
		\end{tabular}
	}
    \vspace{-7pt}
    \caption{\textbf{Depth estimation results on Cityscapes~\cite{cordts2016cityscapes}.} We compare our model with previous single- and multi-frame depth estimation networks. For our baseline, we adopted the CroCo-stereo architecture~\cite{weinzaepfel2023croco} and trained it using a self-supervised depth learning manner. The best is in \textbf{bold}, and the second-best is \underline{underlined}. $\dagger$: our reproduced results from the official repository, $\mathcal{M}$: monocular depth~\citep{godard2019digging}, $\mathcal{S}$: segmentation, and $\mathcal{F}$: flow network.
    }
    \label{suptab:cityscape}
\end{table*}

\clearpage
\section{Additional Qualitative Results}
\label{subsec:qual}
The additional qualitative results on zero-shot matching, as well as learning-based geometric matching and depth estimation, highlight the effectiveness of the Cross-View Completion (CVC) pretext task~\cite{weinzaepfel2023croco}, which enables the cross-attention map to learn geometric relationships that can either be used directly or fine-tuned for optimal performance in downstream tasks.

\subsection{Zero-shot Matching}
\begin{figure*}[!t]
    \centering
    \includegraphics[width=\linewidth]{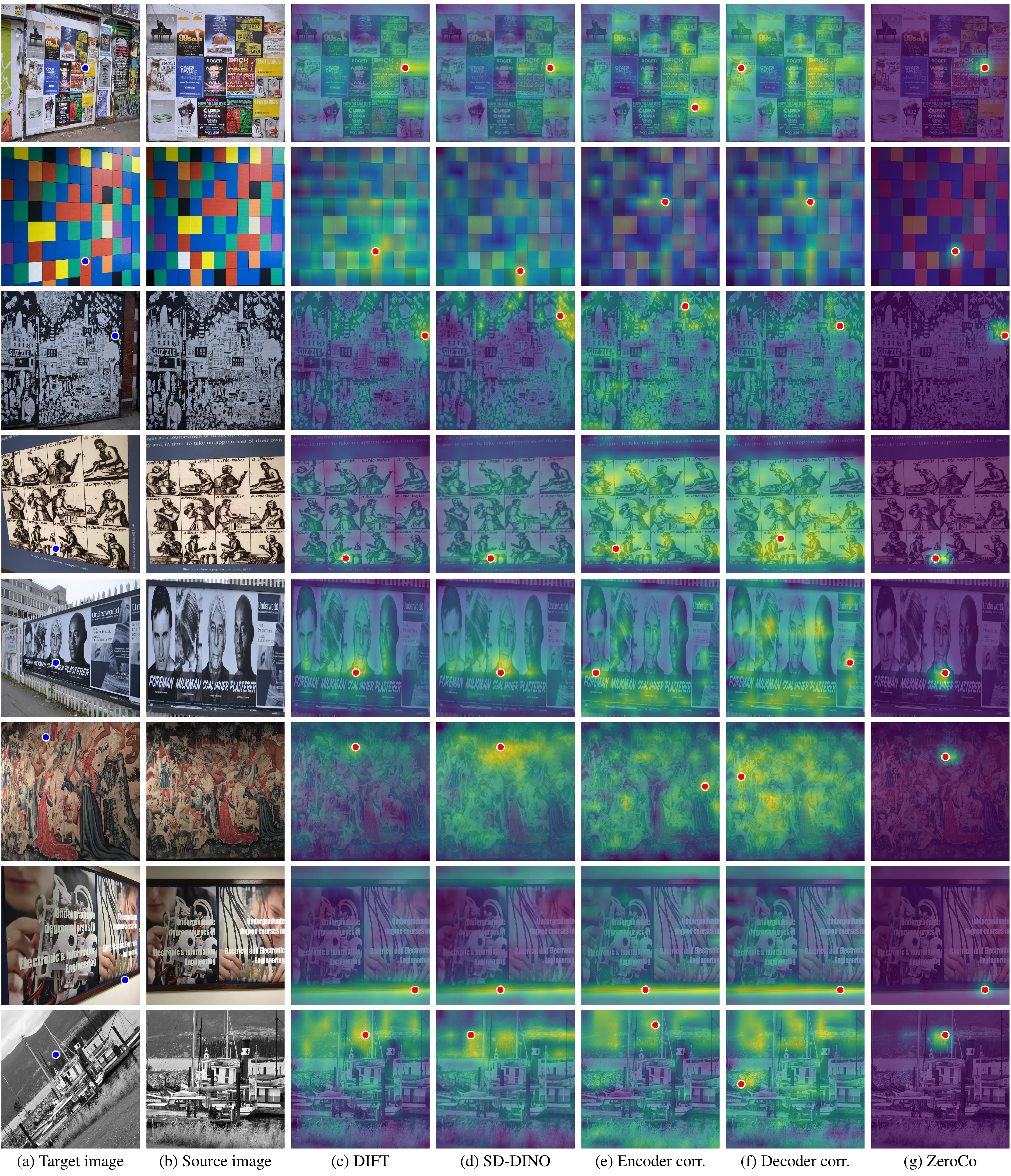}
    \vspace{-15pt}
    \caption{\textbf{Visualization of matching costs in previous zero-shot matching methods~\cite{tang2023emergent,zhang2024tale}, encoder and decoder features within cross-view completion models, and our ZeroCo.}} 
    \vspace{-10pt}    
    \label{supfig:attn1}
\end{figure*}

\begin{figure*}[!t]
    \centering
    \includegraphics[width=\linewidth]{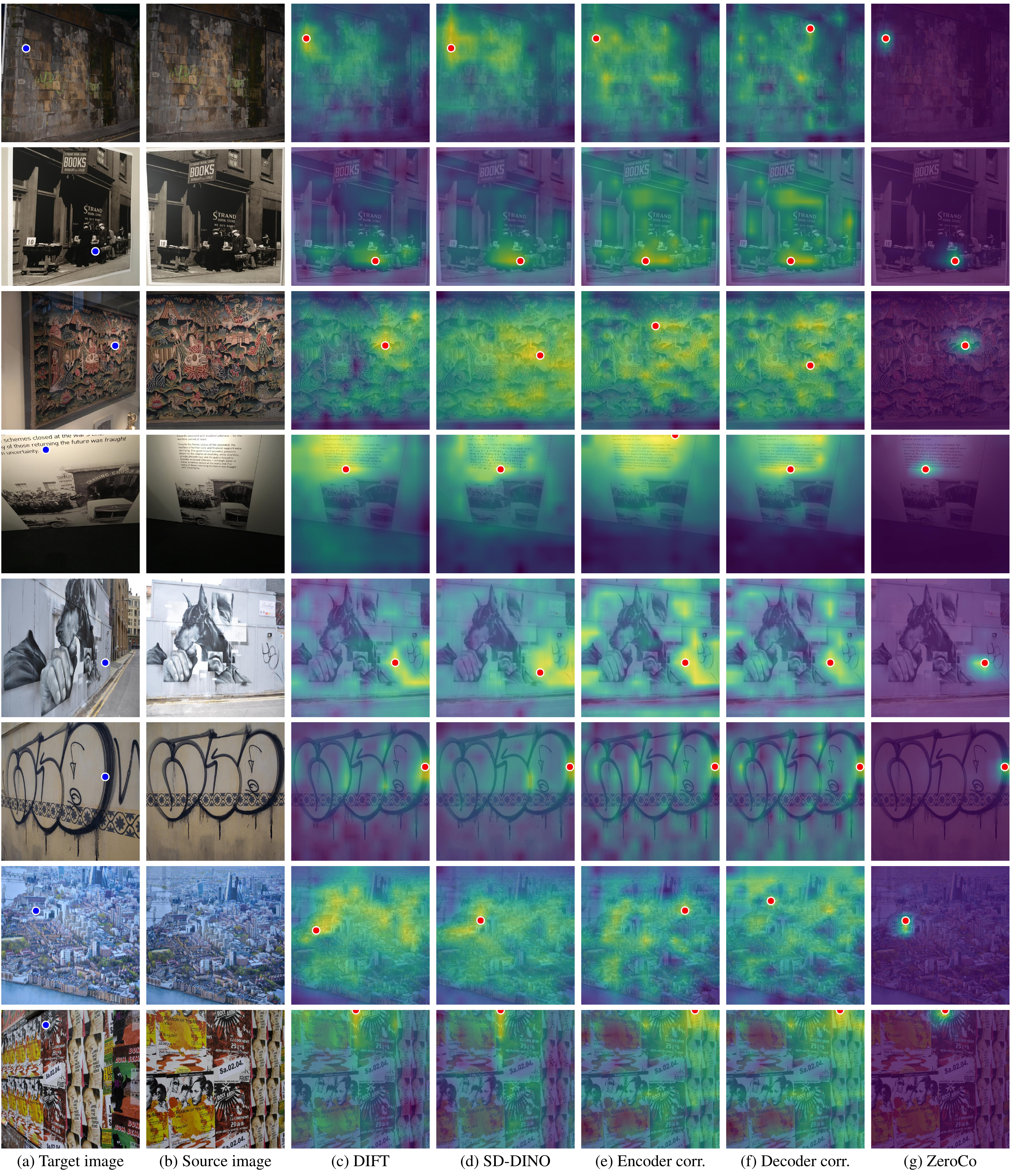}
    \vspace{-15pt}\caption{\textbf{Visualization of matching costs in previous zero-shot matching methods~\cite{tang2023emergent,zhang2024tale}, encoder and decoder features within cross-view completion models, and our ZeroCo.}} 
    \vspace{-10pt}
    
    \label{supfig:attn2}
\end{figure*}

\begin{figure*}[!t]
    \centering
    \includegraphics[width=\linewidth]{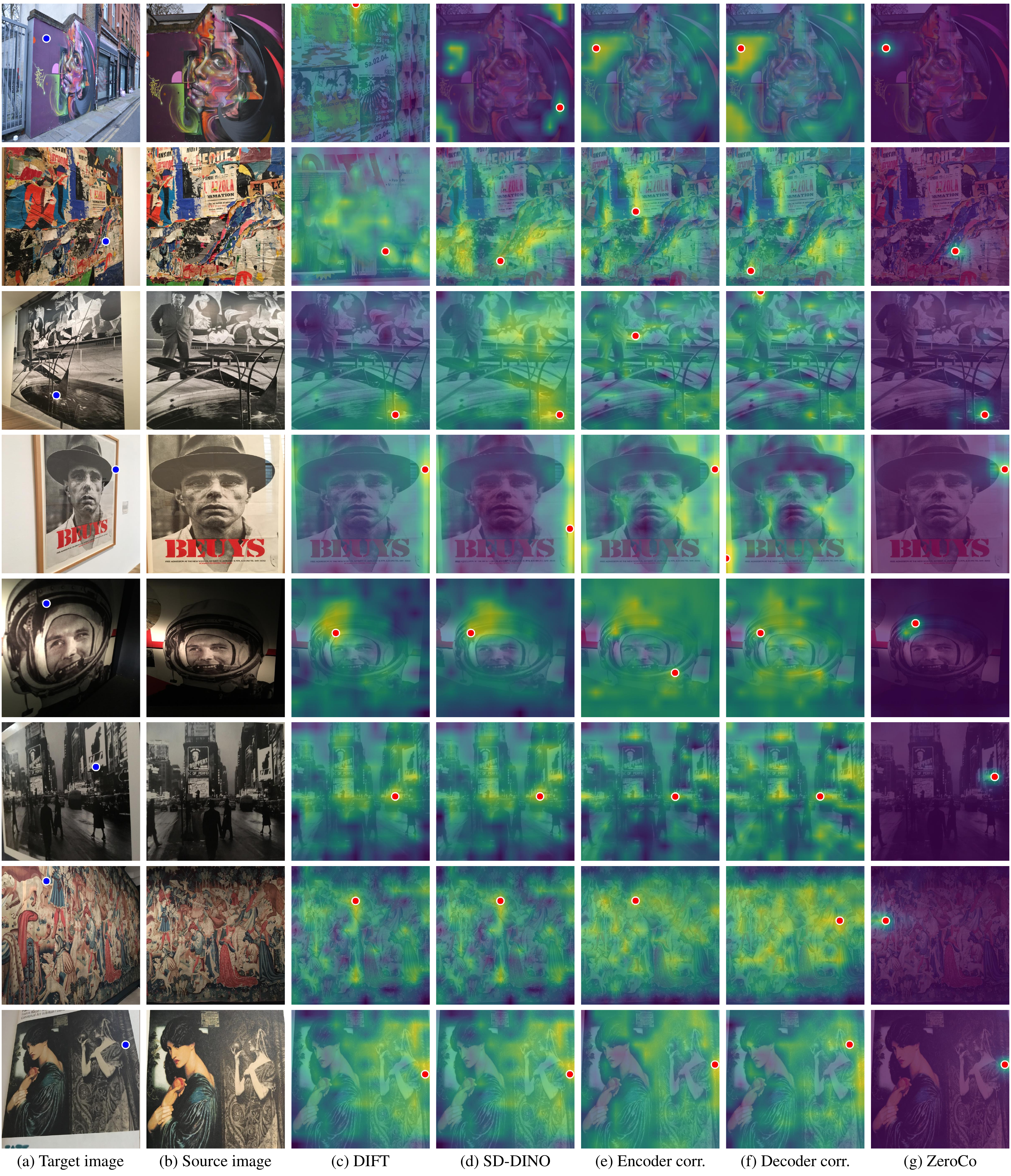}
    \vspace{-15pt}
    \caption{\textbf{Visualization of matching costs in previous zero-shot matching methods~\cite{tang2023emergent,zhang2024tale}, encoder and decoder features within cross-view completion models, and our ZeroCo.}} 
    \vspace{-10pt}
    
    \label{supfig:attn3}
\end{figure*}

Fig.~\ref{supfig:attn1}, \ref{supfig:attn2}, and \ref{supfig:attn3} showcase correlations from encoder and decoder features and cross-attention maps for target and source images. Blue points mark the query point in the target image, with the corresponding red point indicating the highest attention value based on matching costs.

\begin{figure*}[!t]
    \centering
    \includegraphics[width=0.9\linewidth]{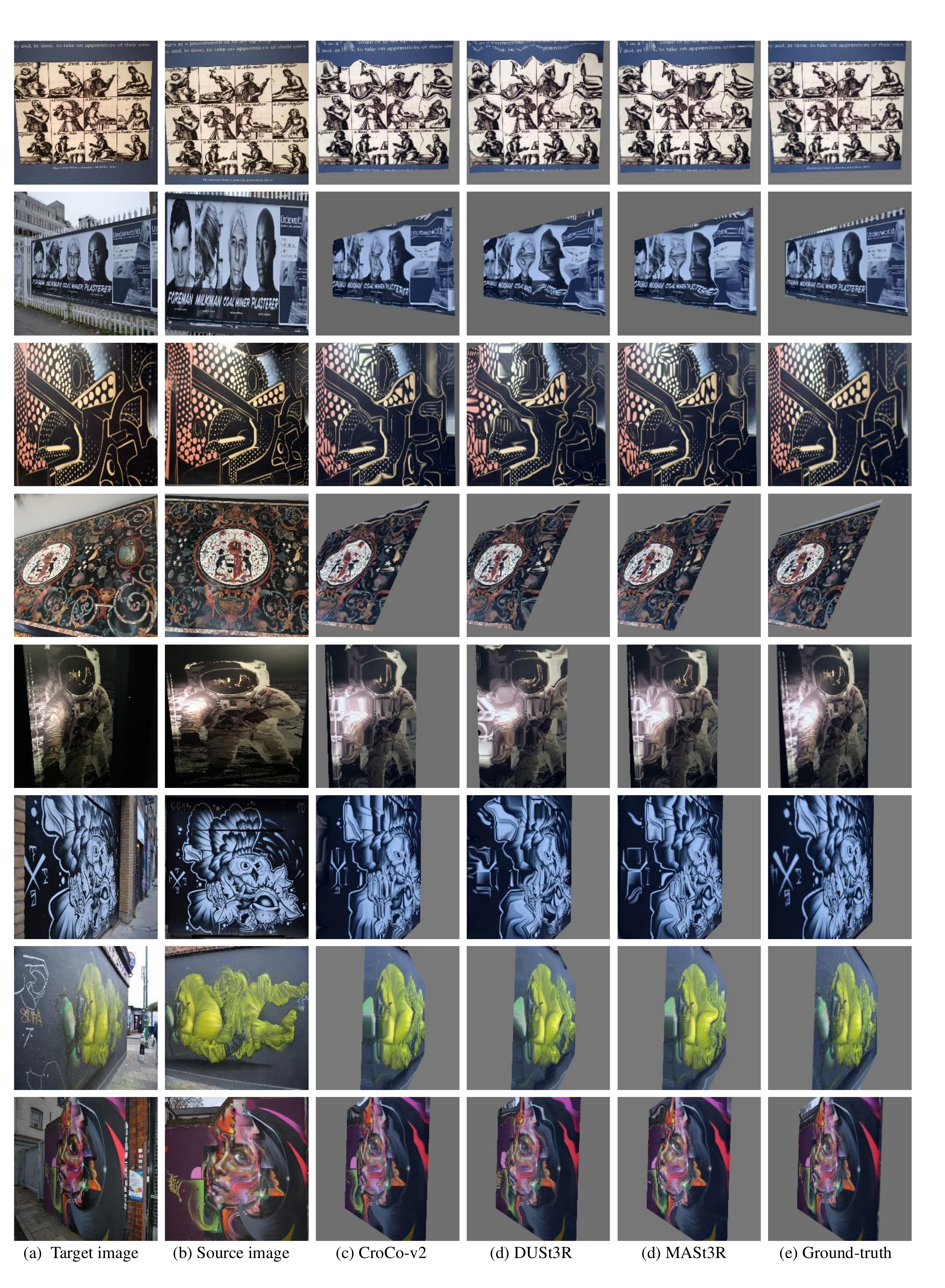}
    \vspace{-15pt}
    \caption{\textbf{Visualization of warped images using cross-attention maps.} Based on our findings, we used the \textbf{cross-attention} maps from CroCo-v2~\cite{weinzaepfel2023croco}, DUSt3R~\cite{wang2024dust3r}, and MASt3R~\cite{leroy2024grounding} to warp the source image to the respective target image, which shows the effectiveness of the cross-attention maps in various cross-view completion-based models for dense correspondence.} 
    \vspace{-10pt}
    
    \label{supfig:vis_warp}
\end{figure*}

\begin{figure*}[!t]
    \centering
    \includegraphics[width=0.9\linewidth]{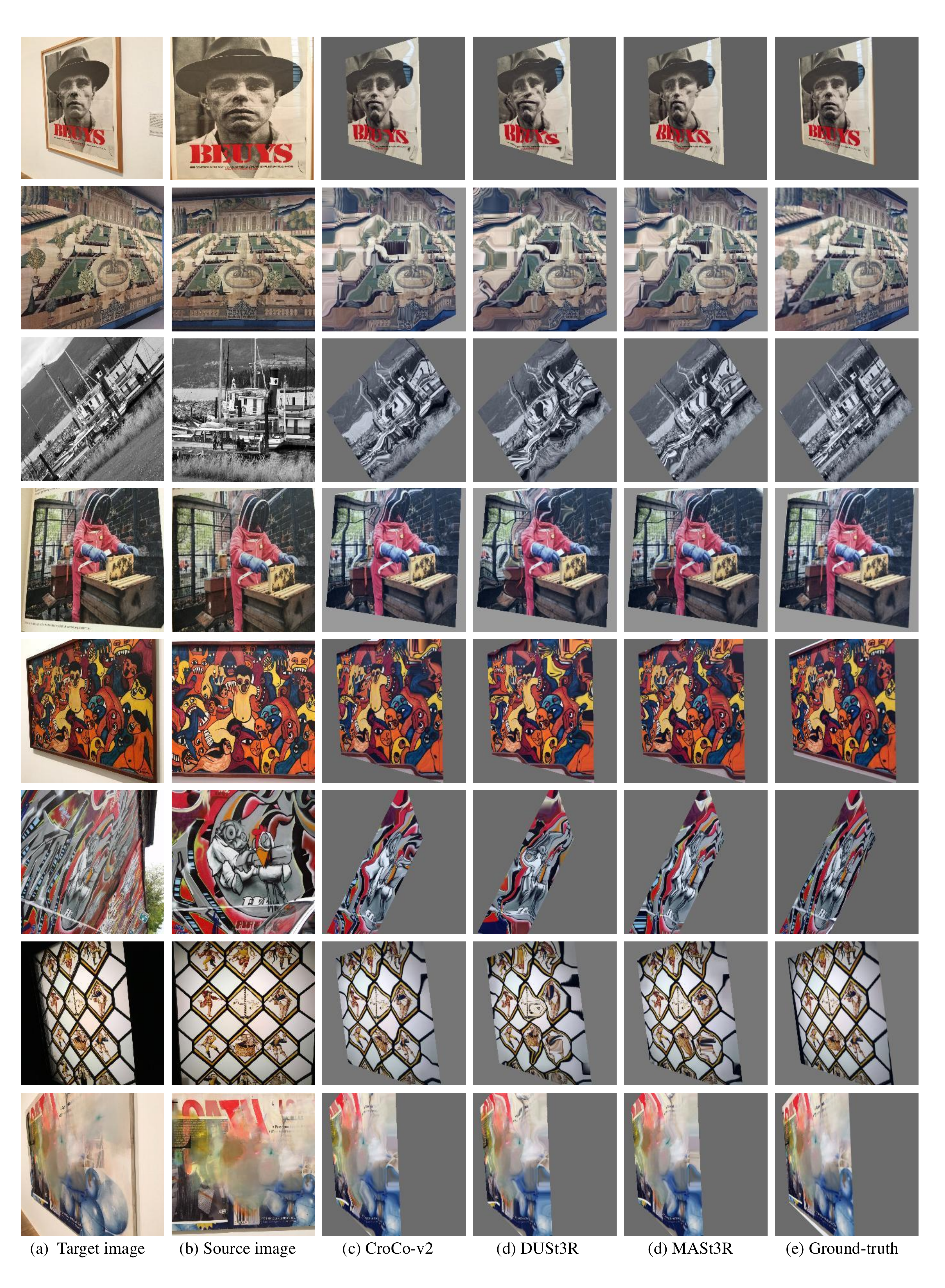}
    \vspace{-15pt}
    \caption{\textbf{Visualization of warped images using cross-attention maps.} Based on our findings, we used the \textbf{cross-attention} maps from CroCo-v2~\cite{weinzaepfel2023croco}, DUSt3R~\cite{wang2024dust3r}, and MASt3R~\cite{leroy2024grounding} to warp the source image to the respective target image, which shows the effectiveness of the cross-attention maps in various cross-view completion-based models for dense correspondence.} 
    \vspace{-10pt}
    
    \label{supfig:vis_warp2}
\end{figure*}

Fig.~\ref{supfig:vis_warp} and \ref{supfig:vis_warp2} present additional visualizations of warped images, leveraging the cross-attention maps from the successful prior works DUSt3R \cite{wang2024dust3r} and MASt3R \cite{leroy2024grounding}. As both models were also pretrained on CVC, their attention maps effectively demonstrate their effectiveness for zero-shot matching.

\subsection{Learning-based Geometric Matching}
\begin{figure*}[!t]
    \centering
    \includegraphics[width=0.9\linewidth]{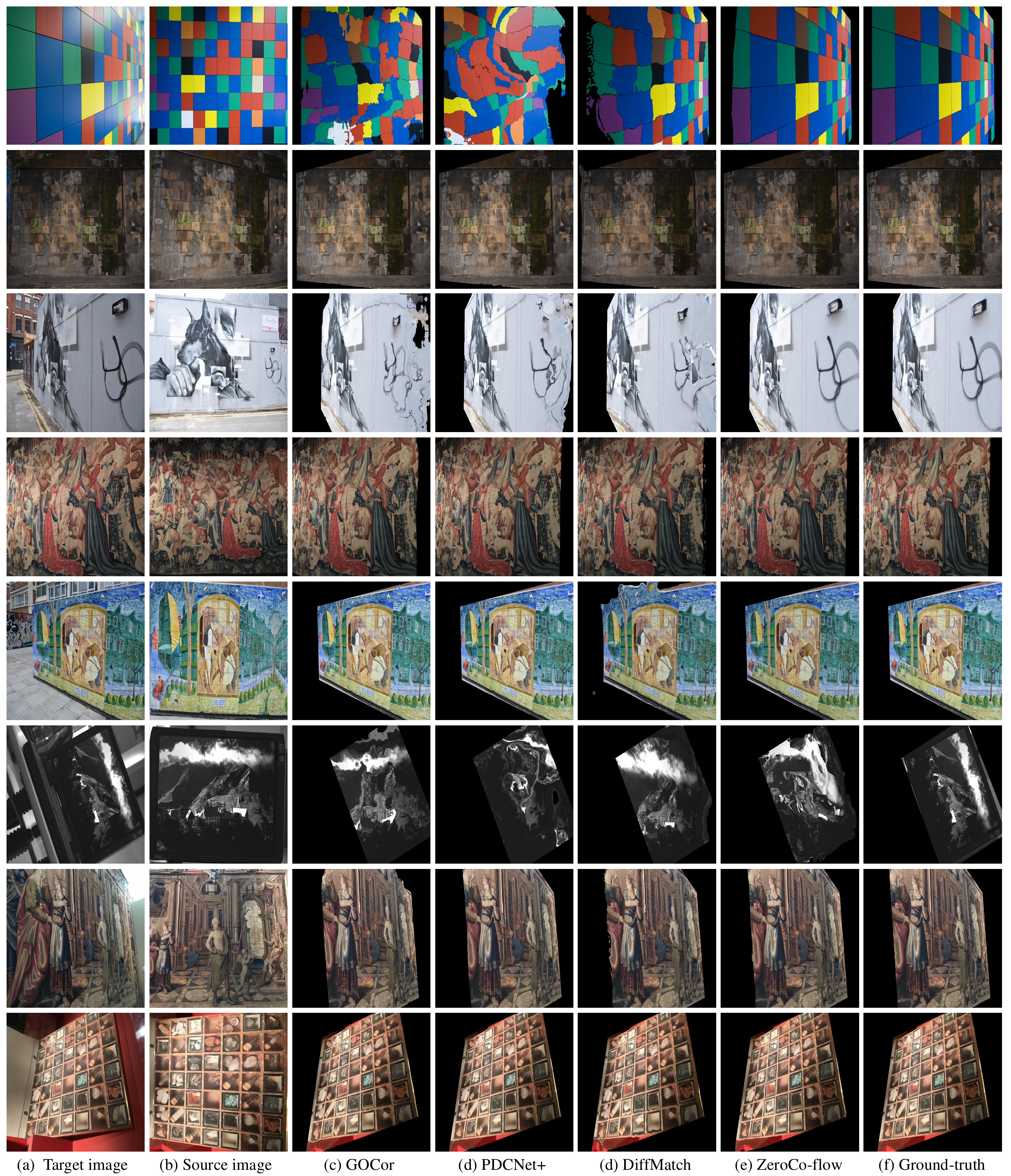}
    \vspace{-5pt}
    \caption{\textbf{Visualization of warped images using estimated dense correspondence.} We used the output flow from GLU-Net-GOCor~\cite{truong2020gocor}, PDCNet+~\cite{truong2023pdc}, DiffMatch~\cite{namdiffusion}, and our ZeroCo-flow to warp the source image to the respective target image.} 
    \vspace{-10pt}
    
    \label{supfig:vis_warp3}
\end{figure*}

Fig.~\ref{supfig:vis_warp3} presents visualizations of warped images using the dense output flow from learning-based geometric matching methods. In comparison to recent dense matching models that were not pretrained with CVC, (e) ZeroCo-flow exhibits superior performance. This is due to the geometric knowledge acquired through CVC, which enables the model to achieve improved matching and warping results, as seen in the first and third rows.

\subsection{Learning-based Multi-frame Depth Estimation}
\begin{figure*}[!t]
    \centering
    \includegraphics[width=0.9\linewidth]{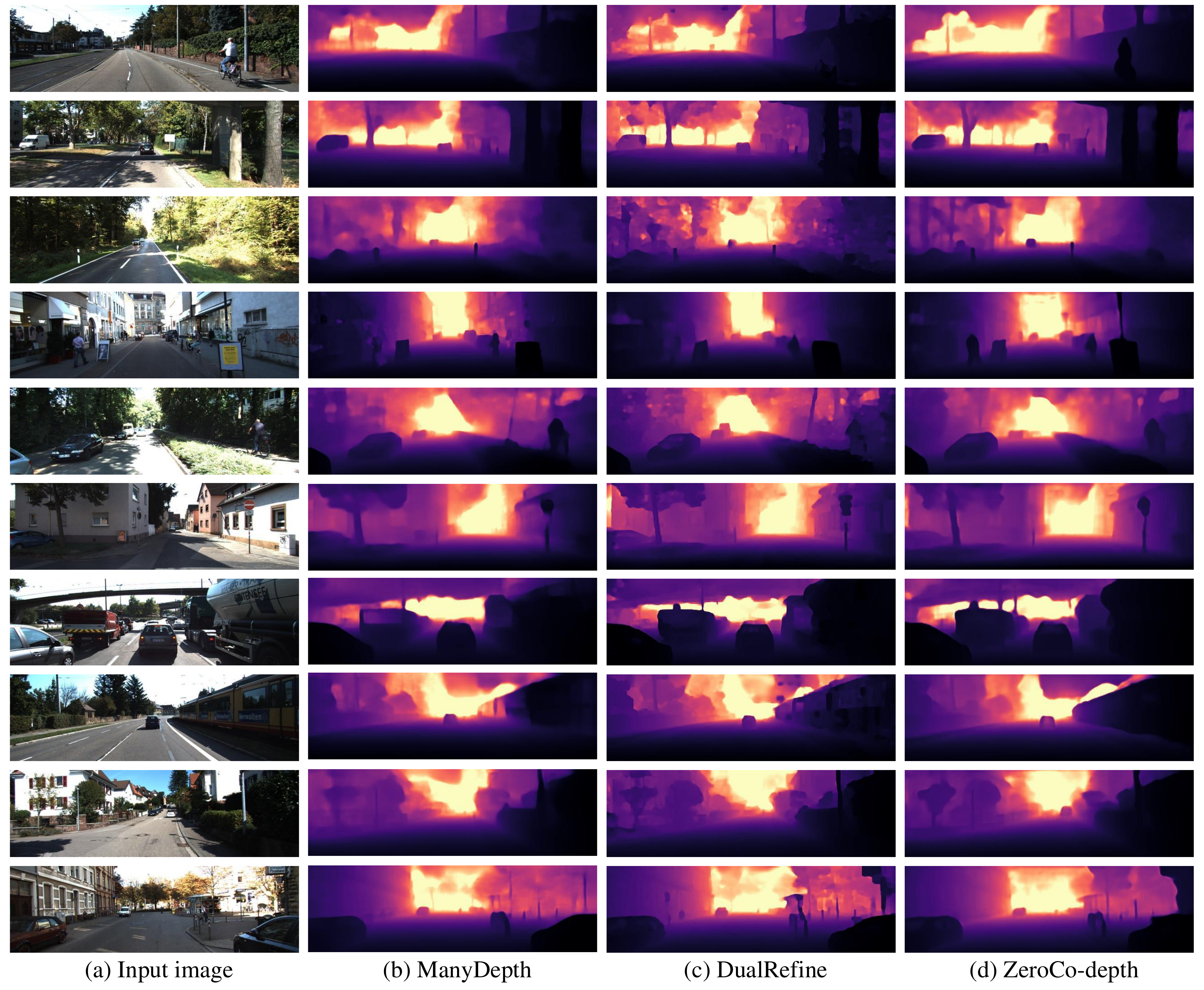}
    \vspace{-4pt}
    \caption{\textbf{Qualitative results for multi-frame depth estimation on the KITTI~\cite{geiger2013vision} dataset.} We compare our ZeroCo-depth with multi-view depth estimation models that leverage epipolar-based cost volumes~\cite{watson2021temporal,bangunharcana2023dualrefine} and demonstrate improved depth prediction performance for dynamic objects through a full cost volume represented by a cross-attention map.} 
    \vspace{-10pt}
    
    \label{supfig:vis_kitti}
\end{figure*}

\begin{figure*}[!t]
    \centering
    \includegraphics[width=0.9\linewidth]{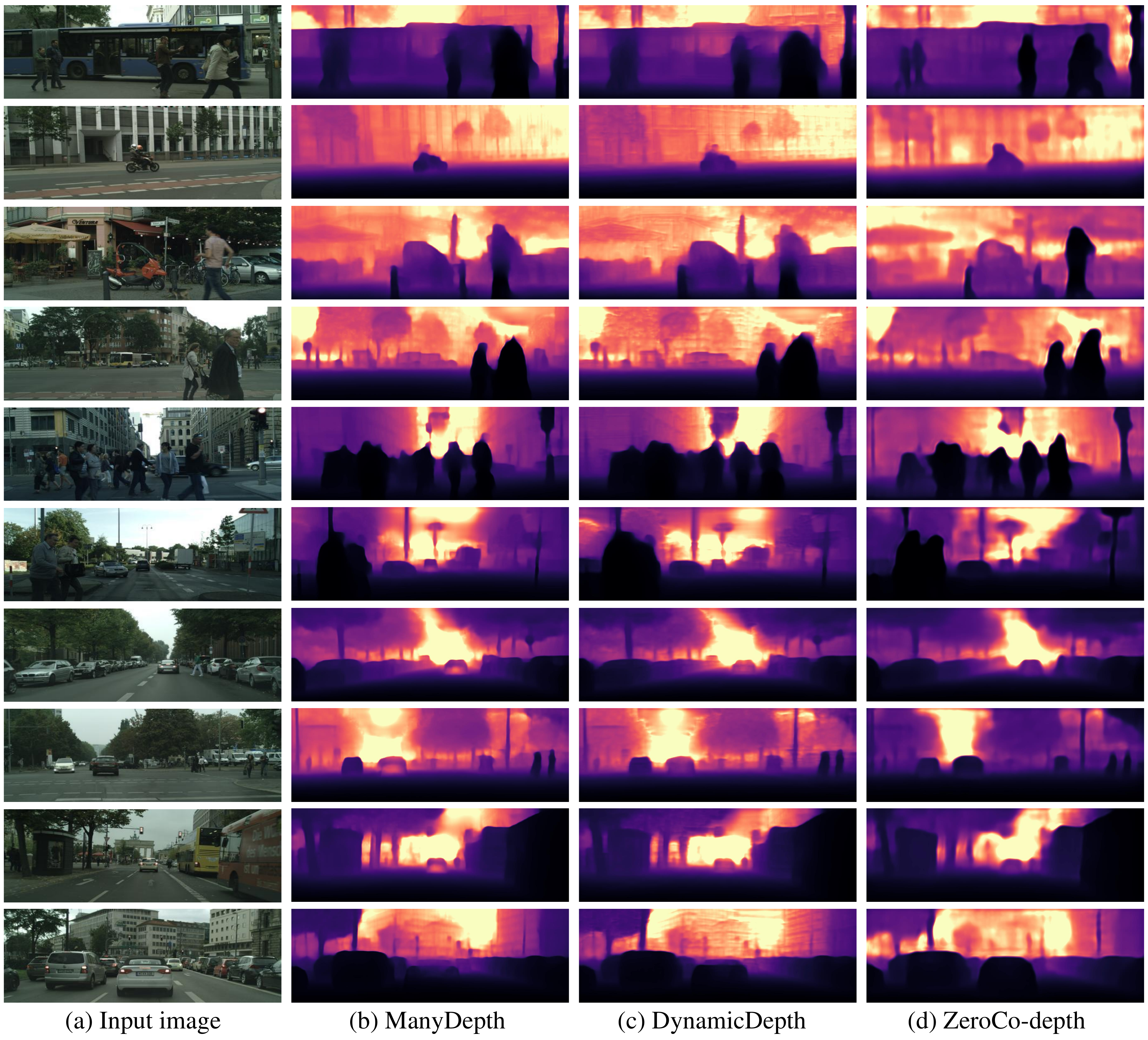}
    \vspace{-3pt}
    \caption{\textbf{Qualitative results for multi-frame depth estimation on the Cityscapes~\cite{cordts2016cityscapes} dataset.} We compare our ZeroCo-depth with multi-view depth estimation models that leverage epipolar-based cost volumes~\cite{watson2021temporal,feng2022disentangling} and demonstrate improved depth prediction performance for dynamic objects through a full cost volume represented by a cross-attention map.} 
    \vspace{-10pt}
    
    \label{supfig:vis_cityscape}
\end{figure*}

Fig.~\ref{supfig:vis_kitti} and \ref{supfig:vis_cityscape} display multi-frame depth estimation results from our model and state-of-the-art models on the KITTI~\cite{geiger2013vision} and Cityscapes~\cite{cordts2016cityscapes} datasets. In contrast to previous multi-frame depth estimation methods that use epipolar-based cost volumes to explicitly model matching costs, our cross-attention map acts as a full cost volume, which more effectively captures geometric relationships and matching costs, allowing for better detection of dynamic objects such as cars and pedestrians.

\clearpage
\section{Limitation}
\label{subsec:limitation}
Our method may face challenges with semantic object correspondence tasks~\cite{kim2018recurrent, cho2021cats, cho2022cats++}, which involve additional complexities such as intra-class variations and background clutter in image pairs. However, this limitation could potentially be addressed by training cross-view completion models on semantically similar object pairs. Additionally, our method is currently constrained to two-view inputs, and extending it to handle a large number of view inputs would require further implementation efforts and considerations. Finally, the method may encounter difficulties when applied to extremely high-resolution images.

\clearpage



\end{document}